\documentclass[letterpaper,twocolumn,10pt]{article}

\usepackage{hegroup}

\usepackage[utf8]{inputenc}
\usepackage[T1]{fontenc}
\usepackage{hyperref}
\usepackage{url}
\usepackage{booktabs}
\usepackage{amsfonts}
\usepackage{nicefrac}
\usepackage{microtype}
\usepackage{xcolor}
\usepackage{graphicx}
\usepackage{multirow} 
\usepackage{subcaption}
\usepackage{siunitx}
\usepackage{threeparttable}
\usepackage{enumitem}
\usepackage{xspace}
\definecolor{nicegreen}{RGB}{34,139,34}
\newcommand{\mypara}[1]{\noindent{\bf {#1}.}\xspace}
\usepackage{xcolor}
\newcommand{\dataset}{\textit{FragFake}\xspace}
\usepackage[most]{tcolorbox}
\usepackage{fontawesome5}
\usepackage{cleveref}
\usepackage{wrapfig}
\usepackage{array}
\usepackage{enumitem}
\newcolumntype{C}[1]{>{\centering\arraybackslash}m{#1}}

\title{\includegraphics[height=1.0em]{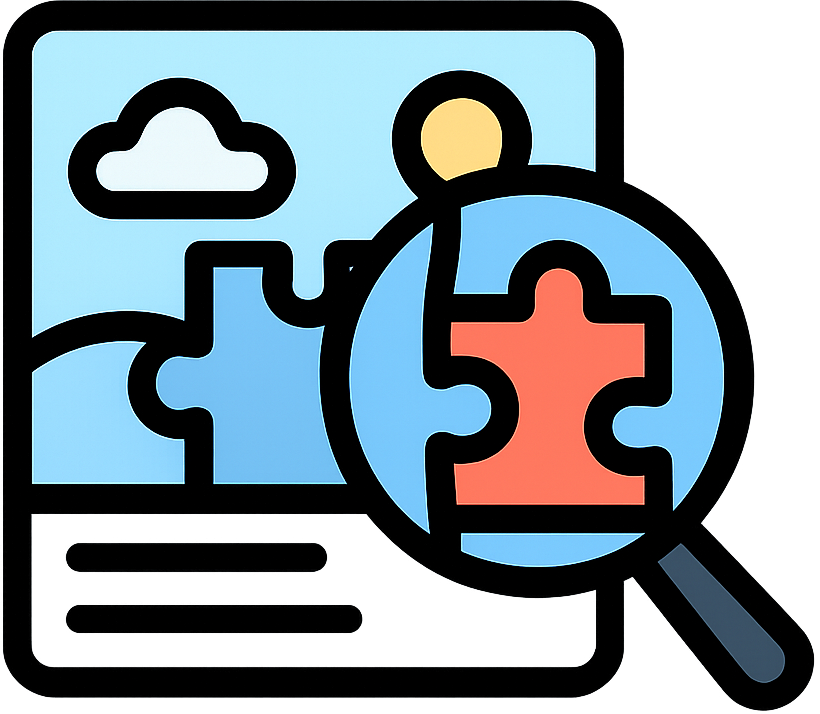}~Can VLMs Detect and Localize Fine-Grained AI-Edited Images?}
\date{}

\author{
\textbf{Zhen Sun}$^{1}$ \quad
\textbf{Ziyi Zhang}$^{1}$ \quad
\textbf{Zeren Luo}$^{1}$ \quad 
\textbf{Zhiyuan Zhong}$^{1}$ \quad 
\textbf{Zeyang Sha}$^{2}$ \quad \\
\textbf{Tianshuo Cong}$^{4}$ \quad
\textbf{Zheng Li}$^{4}$ \quad
\textbf{Shiwen Cui}$^{2}$ \quad
\textbf{Weiqiang Wang}$^{2}$ \quad \\
\textbf{Jiaheng Wei}$^{1}$ \quad
\textbf{Xinlei He}$^{1}$\thanks{Corresponding author(\href{mailto:xinleihe@hkust-gz.edu.cn}{xinleihe@hkust-gz.edu.cn}).} \quad
\textbf{Qi Li}$^{3}$ \quad
\textbf{Qian Wang}$^{5}$ \\
$^1$Hong Kong University of Science and Technology (Guangzhou)
$^2$Ant Group \\
$^3$Tsinghua University
$^4$Shandong University
$^5$Wuhan University
}

\begin{document}

\maketitle

\begin{abstract}
Fine-grained detection and localization of localized image edits is crucial for assessing content authenticity, especially as modern diffusion models and image editors can produce highly realistic manipulations.
However, this problem faces three key challenges: (1) most AIGC detectors produce only a global real-or-fake label without indicating where edits occur;
(2) traditional computer vision methods for edit localization typically rely on costly pixel-level annotations; and
(3) there is no large-scale, modern benchmark specifically targeting edited-image detection. 
To address these gaps, we develop an automated data-generation pipeline and construct \dataset, a large-scale benchmark of AI-edited images spanning multiple source datasets, diverse editing models, and several common edit types.
Building on \dataset, we are the first to systematically study vision language models (VLMs) for edited-image classification and edited-region localization.
Our experiments show that pretrained VLMs, including GPT4o, perform poorly on this task, whereas fine-tuned models such as Qwen2.5-VL achieve high accuracy and substantially higher object precision across all settings.
We further explore GRPO-based RLVR training, which yields modest metric gains while improving the interpretability of model outputs. 
Ablation and transfer analyses reveal how data balancing, training size, LoRA rank, and training domain affect performance, and highlight both the potential and the limitations of cross-editor and cross-dataset generalization.
We anticipate that this work will establish a solid foundation to facilitate and inspire subsequent research endeavors in the domain of multimodal content authenticity.
\end{abstract}

\begin{figure*}
    \centering
    \includegraphics[width=1\linewidth]{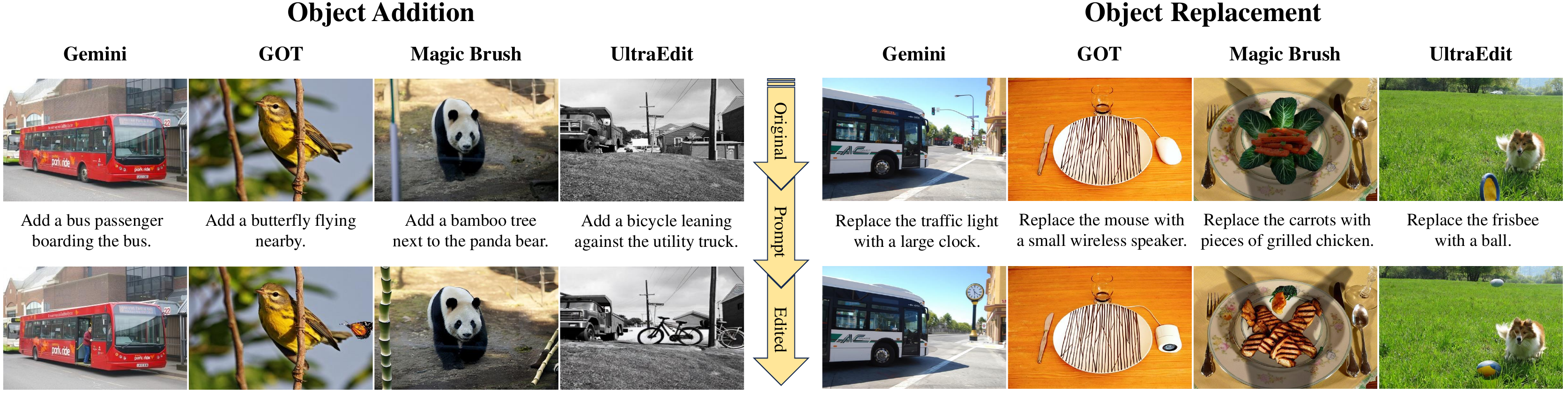}
    \caption{Examples of edited images generated by four different models, showcasing two types of operations: Object Addition and Object Replacement.}
    \label{fig:addition_and_replacement}
\end{figure*}

%-----------------------
\section{Introduction}
%-----------------------

Owing to the swift progress of diffusion models, the images they generate, commonly referred to as AI-Generated Content (AIGC), have become remarkably lifelike~\cite{DBLP:journals/air/ChenXHYYCZ25,DBLP:journals/csur/YangZSHXZZCY24,DBLP:conf/cvpr/RombachBLEO22}.
At the same time, text-guided image editing techniques have also made significant advances~\cite{DBLP:journals/corr/abs-2402-17525,DBLP:conf/cvpr/BrooksHE23,DBLP:conf/nips/ZhangMCSS23}, enabling localized modifications driven by natural language instructions while preserving the rest of the image~\cite{DBLP:journals/corr/abs-2402-17525}.

Compared with fully synthetic images, partially edited images from real photographs are more insidious, as small local edits to largely genuine content can drastically alter how a scene is perceived.
When such content is circulated on online platforms, it can inadvertently fuel image-based misinformation and manipulate public opinion~\cite{DBLP:journals/corr/abs-2404-03021}, or be deliberately exploited by attackers as disinformation to fabricate evidence and cause financial losses to individuals and the broader public.
For example, in July 2024, a genuine Associated Press photograph of Secret Service agents protecting Donald Trump after an assassination attempt was circulated alongside an edited version in which the agents were made to appear smiling, leading some users to view the incident as staged and contributing to political misinformation~\cite{apnews2024}.
In another case, an Airbnb host reportedly submitted AI-edited images of fabricated property damage to support a false \pounds$12,000$ compensation claim against a guest, showing how manipulated images can be used as disinformation to create fake evidence and cause substantial financial harm~\cite{airbnb_guardian}.
Taken together, these cases show that realistic edited images pose serious safety and societal risks, underscoring the need for robust methods to accurately distinguish partially edited images from genuine ones and thereby mitigate image-based misinformation and disinformation on modern online platforms.

Most traditional AIGC detectors are trained on datasets consisting entirely of either real or fully generated images. 
As a result, their performance degrades significantly when faced with images that contain only localized edits.
For example, with open-source AIGC detector Hive Moderation~\cite{hivemoderation}, only 55 out of 100 partially edited images are correctly identified as AI-generated (as described in~\Cref{sec:hive}). 
This limitation arises because a large portion of the pixels remain authentic, which biases the classifier toward predicting the entire image as real.
In addition, most existing detectors adopt a binary classification strategy that produces only an image-level ``real'' or ``fake'' decision, without indicating which specific regions have been edited.
This lack of spatial interpretability restricts their practical use in real-world forensic and provenance applications.
Although some computer vision approaches explore edited-region localization~\cite{DBLP:conf/mm/SunF0ZW24}, they typically require costly pixel-level annotations and are trained on datasets built with outdated editing models that no longer reflect the realism of modern generation techniques.
Nowadays, powerful vision language models (VLMs), pretrained on large-scale image-text corpora, can be efficiently adapted to many downstream tasks with light-weight fine-tuning~\cite{li2025surveystateartlarge}.
This naturally raises the question of whether such models can also be used to detect and localize subtle image edits. 

%-----------------------
\subsection{Our Contribution}
\label{sec:contribution}
%-----------------------

To answer this question, we reframe edited image detection (both image-level classification and edited-region localization) as a vision-language understanding task, aiming to leverage VLMs' multimodal reasoning while reducing reliance on costly pixel-level annotations.

Since using VLMs for edited image detection is a novel task and no high-quality public dataset currently exists, we construct a dedicated image dataset, \dataset.
It consists of images edited by six advanced models: $5$ open-source editors (MagicBrush~\cite{DBLP:conf/nips/ZhangMCSS23}, GoT~\cite{DBLP:conf/nips/ZhaoMCSWAYZLC24}, UltraEdit~\cite{DBLP:conf/nips/ZhaoMCSWAYZLC24}, Flux~\cite{DBLP:journals/corr/abs-2506-15742}, Step1X-Edit~\cite{DBLP:journals/corr/abs-2504-17761}) and $1$ commercially deployed editor, Gemini-IG~\cite{Gemini}.
To ensure diversity, \dataset covers $4$ types of editing operations: object addition and object replacement on COCO~\cite{DBLP:conf/eccv/LinMBHPRDZ14} and ADE20K~\cite{zhou2017scene}, background change on ADE20K, and facial expression change on FFHQ~\cite{10.1109/TPAMI.2020.2970919}.

Since most modern image editors support natural language-driven editing, we use GPT4o~\cite{openai2024gpt4o} to generate editing instructions based on these original images. 
During this process, we observe that many target objects specified in the instructions are repeated. 
We therefore refer to this subset as the Unfiltered (UF) split.
Building upon it, we further refine the dataset by filtering and replacing overlapping target objects to produce the Unique (UQ) split.
Combined with $6$ editors, these instructions produce $98,412$ edited images.
Both image and instruction generation are fully automated, enabling scalability.
The resulting edited images and their associated instructions are converted into image-text pairs for training VLMs, and we fine-tune four widely used models for this task: LLaVA-1.5~\cite{DBLP:conf/nips/LiuLWL23a}, Qwen2-VL~\cite{DBLP:journals/corr/abs-2409-12191}, Qwen2.5-VL~\cite{DBLP:journals/corr/abs-2502-13923}, and Gemma3~\cite{DBLP:journals/corr/abs-2503-19786}.

Our evaluation operates at two levels: image-level edited-image classification, measured by accuracy (Acc) and F1-score, and edited-region localization, measured by Region Precision (RP) and Object Precision (OP). On the COCO subset generated by Gemini-IG, the best pretrained VLM, GPT4o, reaches only around $0.81$-$0.83$ Acc and $45$-$46\%$ OP, while several other pretrained models achieve OP scores close to random guessing.
After fine-tuning on \dataset, Qwen2.5-VL becomes the strongest detector, attaining close to $0.99$ Acc with OP in the $70\%+$ range on the same splits, and showing similarly strong performance on ADE20K and on additional editing types such as background change and facial expression change.
These results indicate that large pretrained VLMs alone are far from solving fine-grained edit detection, but once adapted on \dataset they can serve as highly accurate and fine-grained detectors.

Beyond this main comparison, we conduct comprehensive analyses to understand what drives performance.
Ablation studies show that simple data balancing, training sets, and appropriately chosen LoRA ranks all yield steady gains, while GRPO-based RLVR training~\cite{DBLP:journals/corr/abs-2507-04136} offers modest improvements and more interpretable outputs.
Transfer experiments across editors, datasets, and editing tasks reveal that single-source training leads to localization drops on unseen domains, and a user study confirms that humans remain far behind our fine-tuned VLMs in both detection and localization.
In conclusion, our main contributions are as follows:

\begin{itemize}[left=2pt]
    \item We are the first to propose reframing edited image detection (classification and edited region localization) as a vision-language understanding task to reduce reliance on costly annotations. 
    To support this perspective, we construct \dataset, a large-scale benchmark of AI-edited images generated via a fully automated pipeline with multiple editing operations, diverse editing models, and several source image datasets.
    \item We adapt several VLMs to this task using supervised fine-tuning, and further explore GRPO-based RLVR training.
    Our experiments show that training on the \dataset leads to substantial performance gains for all VLMs on fine-grained edit detection.
    \item We provide a comprehensive empirical analysis, including ablations on data balancing, training size, and LoRA rank, transfer studies across editors, original-image datasets, and editing tasks. In addition, a user study shows that non-expert humans lag far behind our fine-tuned detectors in both accuracy and localization, highlighting the practical value and remaining challenges of VLM-based edited-image detection.
\end{itemize}

\begin{figure*}[ht]
    \centering
    \includegraphics[width=1\linewidth]{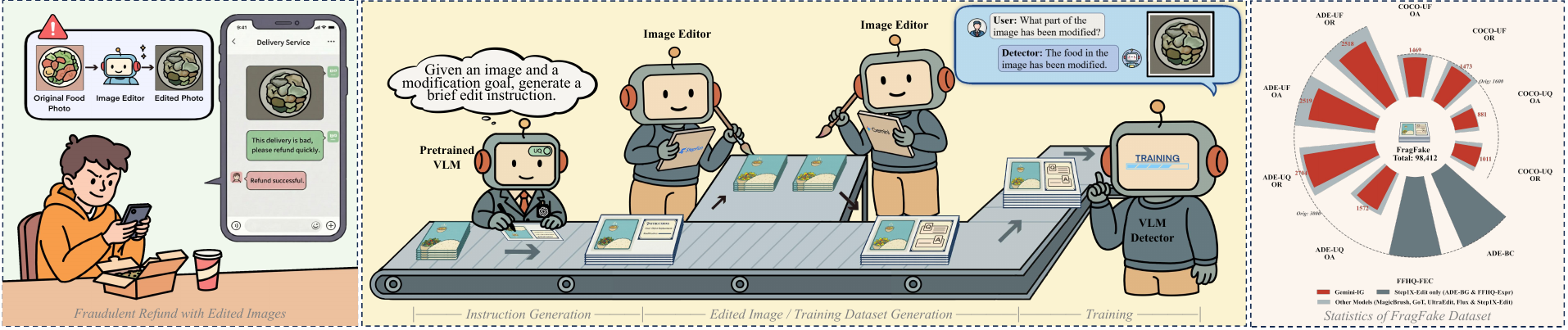}
    \caption{Dataset construction pipeline and \dataset dataset statistics. The left panel shows a real-world fraudulent refund case using edited images; the middle panels depict the instruction generation, edited-image/training data creation, and detector training pipeline; the right panel presents statistics of the \dataset dataset (OA: Object Addition, OR: Object Replacement, BC: Background Change, FEC: Facial Expression Change).}
    \label{tab:dataset_counts}
\end{figure*}

%-----------------------
\section{Related Work}
\label{sec:bg}
%-----------------------

%-----------------------
\subsection{Image Editing}
%-----------------------

Recently, image editing techniques have significantly evolved, enabling users to intuitively modify images by selectively editing specific regions~\cite{DBLP:conf/cvpr/BrooksHE23,DBLP:conf/nips/ZhangMCSS23,DBLP:conf/nips/ZhaoMCSWAYZLC24}. This differs from traditional image generation, as it demands understanding user intent and preserving original image semantics. However, the ease of creating realistic edited images has also increased misuse, including misinformation, fraud, and defamation, highlighting the urgent need for effective detection methods~\cite{DBLP:journals/corr/abs-2404-03021}. 
This work focuses on two main editing techniques: diffusion model-based and closed-source model editing.

\begin{itemize}[left=2pt]
    \item \mypara{Diffusion Model-Based Editing}
    Diffusion models have greatly advanced image editing. 
    MagicBrush fine-tunes InstructPix2Pix on a large-scale annotated dataset, significantly improving image quality~\cite{DBLP:conf/nips/ZhangMCSS23,DBLP:conf/cvpr/BrooksHE23}. UltraEdit automatically generates extensive editing instructions using large language models (LLMs) and real images, enhancing dataset diversity~\cite{DBLP:conf/nips/ZhaoMCSWAYZLC24}. 
    GoT integrates reasoning-guided language analysis with diffusion models to enhance semantic and spatial coherence in edited outputs, demonstrating superior performance~\cite{DBLP:journals/corr/abs-2503-10639}.
    \item \mypara{Closed-Source Model Editing}
    Closed-source models, such as Google's Gemini-IG~\cite{Gemini} and Flux AI's Magic Edit~\cite{MagicEdit}, provide advanced multimodal image generation and editing capabilities. Gemini-IG supports multimodal input and sophisticated editing tasks, while Magic Edit excels at interactive, chat-based editing. However, limited API access restricts their broader usage.
\end{itemize}

%-----------------------
\subsection{Fake Image Detection and Edited Region Localization}
%-----------------------

The proliferation of AI-generated content, particularly realistic manipulated images, has intensified misinformation risks~\cite{DBLP:journals/corr/abs-2412-18148,DBLP:journals/mta/Zhang22a}. 
DE-FAKE integrates detection and attribution models to differentiate between real and fake images~\cite{DBLP:conf/ccs/ShaLYZ23}. 
Systematic evaluations highlight that both humans and automated tools can effectively identify AI-generated images~\cite{DBLP:conf/ccs/HaPBSSZZ24}, but traditional binary classifiers struggle with subtle edits. 
To address this, zero-shot approaches like ZeroFake leverage stability differences during image inversion~\cite{DBLP:conf/ccs/ShaTL0024}. 
Although binary classification methods perform well, fine-grained detection is more important for edited images. 
Prior work, such as~\cite{DBLP:conf/mm/SunF0ZW24}, trains segmentation models using pixel-level annotations, often automated with SAM~\cite{DBLP:journals/corr/abs-2304-02643}, but still incurs high resource costs. 
To reduce this burden, we replace pixel-level masks with VLM-based inference of edited regions and objects, significantly lowering annotation overhead.

%-----------------------
\section{Dataset Construction and Training}
\label{sec:construction}
%-----------------------

In this section, we describe the construction goals and pipeline of our edited image detection dataset \dataset. 

%-----------------------
\subsection{Construction Goals}
%-----------------------

As stated in~\Cref{sec:contribution}, we first need to construct the dataset that can be used to train and evaluate the performance of edited image detection.
The built dataset should have the following properties:
\begin{itemize}[left=2pt]
    \item \textbf{Diversity of Editing Models:}
    We include $6$ image editing models: $1$ closed-source commercial model (Gemini-IG~\cite{Gemini}) and $5$ open-source models (MagicBrush~\cite{DBLP:conf/nips/ZhangMCSS23}, GoT~\cite{DBLP:conf/nips/ZhaoMCSWAYZLC24}, UltraEdit~\cite{DBLP:conf/nips/ZhaoMCSWAYZLC24}, Flux~\cite{DBLP:journals/corr/abs-2506-15742} and Step1X-Edit~\cite{DBLP:journals/corr/abs-2504-17761}).
    This breadth enables robust detector generalization across diverse editing paradigms.
    \item \textbf{Quality:}
    All $6$ models are capable of generating highly realistic edited images, as illustrated in \Cref{fig:addition_and_replacement}.
    To further ensure the reliability of the evaluation, we manually inspect and curate 100 representative test samples for each of the 26 subsets, resulting in $2,600$ human-verified images where the applied edits are correct and unambiguous.
    \item \textbf{Diversity of Edited Objects:}
    To construct broad editing scenarios, we employ GPT4o to generate a wide range of editing instructions (\Cref{sec:pipeline}).
    To eliminate repetition of target objects and increase the challenge of the task, we apply filtering and re-query steps to construct the Unique (UQ) split, in which every edited target object appears only once.
\end{itemize}

%-----------------------
\subsection{Construction Pipeline}
\label{sec:pipeline}
%-----------------------

\mypara{Original Image Datasets}
To build a comprehensive dataset, we start from the widely used COCO dataset~\cite{DBLP:conf/eccv/LinMBHPRDZ14}, randomly sampling 20 images from each category ($1,600$ images in total) as our base.
To further increase scene diversity, we additionally sample $3,000$ high-resolution scene images from ADE20K~\cite{zhou2017scene} and $3,000$ face images from the high-resolution Flickr-Faces-HQ (FFHQ) dataset~\cite{10.1109/TPAMI.2020.2970919}, which allows us to specifically model fine-grained facial expression edits.
The overall data generation pipeline and the resulting dataset statistics are summarized in \Cref{tab:dataset_counts}.

\mypara{Editing Instruction Creation}
These image editing models operate via natural language instructions.
Manually writing these instructions is both time-consuming and labor-intensive, so we use the pretrained VLM GPT4o-2024-11-20 (temperature set to 1) to generate them automatically. 
First, we apply a unified task template (refer to~\Cref{sec:creation_template}) to produce initial editing prompts for all original images. 
Analysis shows that many target objects to be added or used to replace existing ones are repeated.
We refer to this initial collection as the \textbf{Unfiltered (UF)} split. 
To reduce redundancy, we implement a target-object cache: if a newly generated instruction's target object is already in the cache, we append the prompt ``Important: Please do NOT use the following object: [object]'' and query GPT4o to regenerate the instruction.
If the same object still recurs after three attempts, we discard this instruction. 
The remaining instructions and images constitute the \textbf{Unique (UQ)} split, in which every target object appears only once.
\Cref{sec:target} presents the statistics of the target objects.

\mypara{\dataset}
After generating the editing instructions, we apply six editing models to the source images: $5$ open-source models (MagicBrush, UltraEdit, GoT, Flux, and Step1X-Edit), which we run locally in our environment, and $1$ closed-source commercial service, Gemini-IG (also referred to as Gemini-2.0-flash-exp), whose inputs and outputs are subject to built-in content filters.
This filtering blocks some edits, so Gemini-IG produces slightly fewer edited images than the other models, as shown in~\Cref{tab:dataset_counts}.

We consider $4$ types of editing operations: Object Addition (OA), Object Replacement (OR), Background Change (BC), and Facial Expression Change (FEC).
For OA and OR, we generate edited images on both COCO and ADE20K using all six editing models, whereas BC is created only on ADE20K with Step1X-Edit and FEC only on FFHQ with Step1X-Edit.
In total, \dataset contains $98,412$ edited images across all tasks and datasets.
Once all edited images are obtained, we convert them into the image-text pair format required for VLM training. 
Each pair consists of an edited image and a corresponding model response that explicitly identifies the edited object.
All such pairs together form the complete \dataset.

%-----------------------
\subsection{Training}
%-----------------------
We adopt two training paradigms for our VLM: supervised fine-tuning (SFT) and Reinforcement Learning with Verifiable Rewards (RLVR)~\cite{DBLP:journals/corr/abs-2507-04136}.
SFT is the standard choice for VLMs and is relatively efficient in terms of computation.
RLVR is a more recent reinforcement learning approach that optimizes the model against automatically verifiable rewards and can improve both performance and interpretability, but it is substantially more expensive.
In this work, we therefore explore GRPO-based RLVR training only as an additional experiment on this task.
For RLVR, we design a reward function with three components: output format, binary classification, and object localization.
We first require the model to follow a fixed response template \texttt{<think>...</think> + boxed\{...\}} and check this using a regular expression, where the format score contributes $0.1$ to the final reward.
We then use Qwen3-4B~\cite{qwen3_4b_instruct_2507_2025} as an automatic judge that compares the model output with the ground truth and evaluates (i) whether the binary prediction (real or edited) is correct (CLS, weight $0.6$) and (ii) whether the predicted edited object or region matches the ground truth (OBJ, weight $0.3$).

%-----------------------
\section{Experimental Settings}
%-----------------------

\mypara{Evaluation Metrics}
We evaluate edited-image detection at two levels.
At the image level, we treat detection as a binary classification task and report \textit{Accuracy (Acc)} and \textit{F1-score}.
These metrics are computed fully automatically by parsing the model output and performing keyword-based matching.
At a finer granularity, we introduce two localization-oriented metrics: \textit{Region Precision} (RP) and \textit{Object Precision} (OP).
OP measures whether the model correctly identifies the edited object in a semantically accurate way.
To keep this metric objective, we use a pretrained VLM (Qwen3-4B) as an automatic judge that compares the predicted object description with the ground-truth text and decides whether they are semantically equivalent; OP is then computed directly from these VLM-based decisions without human intervention.
RP evaluates whether the predicted edited region spatially aligns with the ground-truth location.
In practice, human annotators are shown the original image, the edited image, the ground-truth description, and the model output, and they determine whether the predicted object lies in the same region as the ground truth, so RP can be judged as correct even when the VLM's textual description differs from the ground truth.
This makes RP a more permissive, region-level criterion than OP and can be interpreted as an upper bound on the achievable localization performance of detectors.
\Cref{fig:Ground_Truth} provides an example of such human evaluation.
All human annotations are performed in Label Studio (\Cref{fig:label_studio}) and are cross-checked by two authors.

\mypara{VLMs for Training}
We fine-tune four VLMs in our experiments, including LLaVA-1.5 (llava-1.5-7b)~\cite{llava15}, Qwen2-VL-7B~\cite{qwen2}, Qwen2.5-VL-7B~\cite{qwen25}, and Gemma3 (gemma-3-4b-it)~\cite{gemma3}.

\mypara{VLMs for Testing}
In addition to the four open-source models described above, we also evaluate $4$ strong commercial closed-source VLMs: GPT4o-mini (2024-07-18)~\cite{openai2024gpt4omini}, GPT4o (2024-11-20)~\cite{openai2024gpt4o}, GLM-4V (glm-4v-plus-0111)~\cite{zhipu2024glm4v}, and Gemini-2.5 (gemini-2.5-flash-preview-04-17)~\cite{google2024gemini25flash}.
All models are accessed via their official APIs with the temperature set to 0.1.

\mypara{Hyperparameters}
We adopt LoRA~\cite{DBLP:conf/iclr/HuSWALWWC22} for SFT on a single NVIDIA L20 GPU for training.
Unless otherwise specified, we set the LoRA rank to 64, the learning rate to 5e-4, the number of training epochs to 5, and the batch size to 16.
We use the final checkpoint after training for evaluation.
For GRPO-based RLVR training, we instead use $8$ NVIDIA H100 GPUs and train for 20 epochs.
Since the UQ and UF splits of \dataset contain different numbers of samples, we control for training size by randomly sampling 3,000 image pairs (edited image and corresponding original image in a 1:1 ratio) from each subset.
When originals are fewer than edited images, we add non-overlapping COCO images to keep a balanced dataset.

\begin{figure}[h!]
    \centering
    \includegraphics[width=0.8\linewidth]{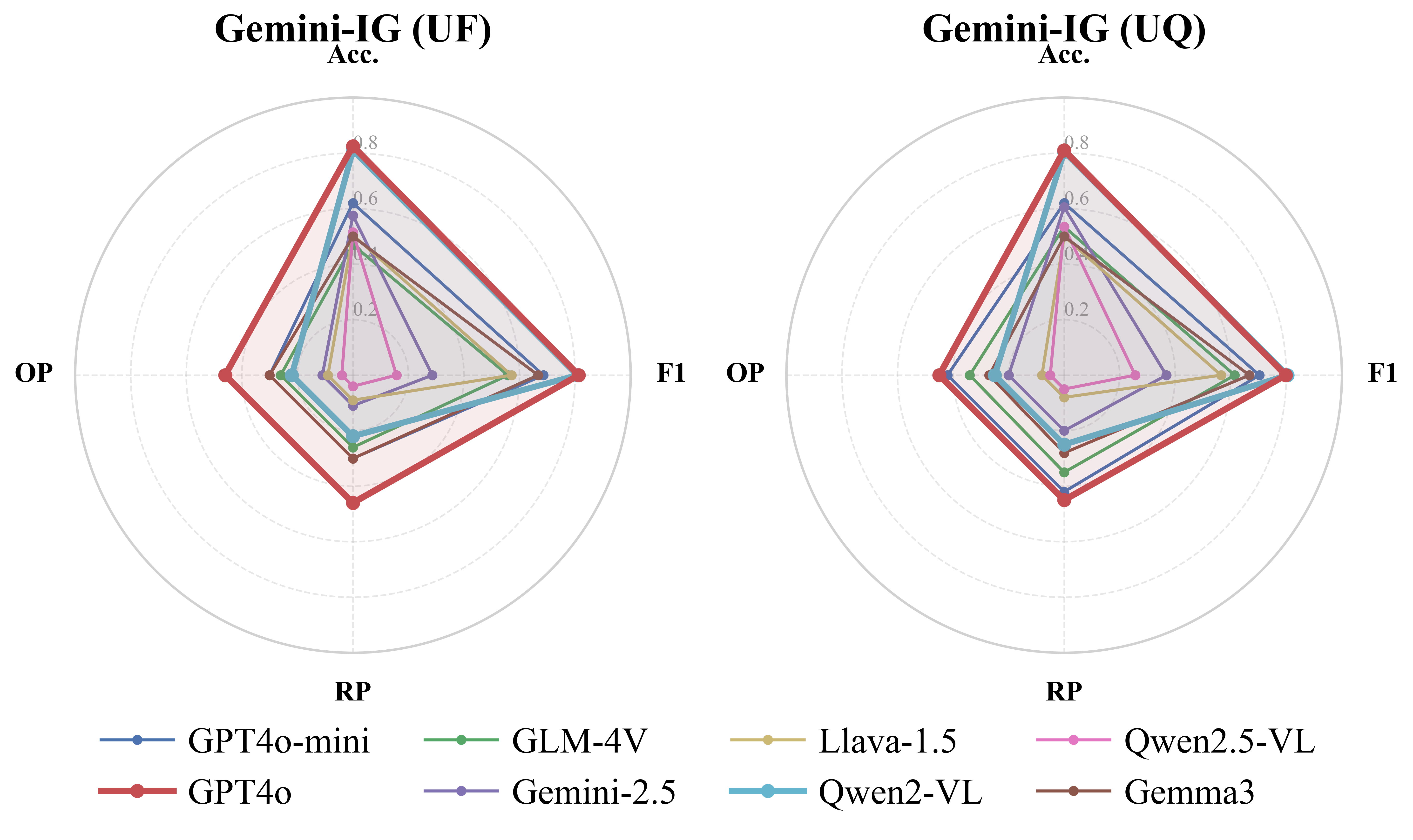}
    \caption{Performance of pretrained VLMs on Gemini-IG.}
    \label{tab:vlm-easy-version-gemini}
\end{figure}

\begin{figure*}[t]
    \centering

    \begin{subfigure}[b]{\linewidth}
        \centering
        \includegraphics[width=0.96\linewidth]{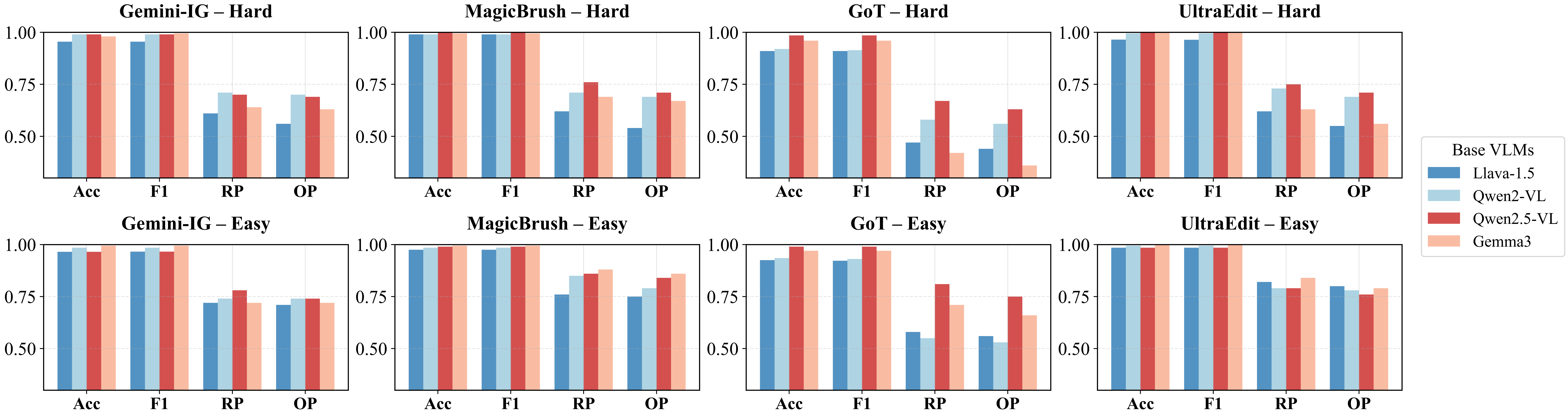}
        \caption{Results on COCO.}
        \label{fig:performance_coco}
    \end{subfigure}
    
    \begin{subfigure}[b]{\linewidth}
        \centering
        \includegraphics[width=0.96\linewidth]{pic/ADE20K_comparison_chart_ADE.png}
        \caption{Results on ADE20K.}
        \label{fig:performance_ade}
    \end{subfigure}

    \caption{Performance comparison of different detectors on UQ and UF splits based on COCO and ADE20K datasets.}
    \label{fig:performance_overall}
\end{figure*}

%-----------------------
\section{Evaluation}
%-----------------------

%-----------------------
\subsection{Comparison of Different VLMs}
%-----------------------

\mypara{Performance of Pretrained VLMs}
VLMs with strong image understanding capabilities often perform well on downstream visual question answering tasks even without fine-tuning. 
To investigate their ability to directly identify whether an object in an image has been edited, we evaluate them on the Gemini-IG subset of the \dataset test set. 
We test two categories of models: (1) popular proprietary production VLMs, including GPT4o-mini, GPT4o, GLM-4V, and Gemini-2.5; (2) widely used open-source VLMs, including Llava-1.5, Qwen2-VL, Qwen2.5-VL, and Gemma3.
For this detection task, we design a unified prompt as demonstrated in~\Cref{sec:edited_template}.

As shown in~\Cref{tab:vlm-easy-version-gemini}, GPT4o achieves the best performance among all detectors, reaching an Acc of $0.825$ and an OP of $46.0\%$ on the UF split of Gemini-IG, and an Acc of $0.810 $ with an OP of $45.0\%$ on the UQ split. 
GPT4o-mini also performs relatively well with an OP of $42.0\%$ on the UQ split but exhibits weaker binary classification performance (Acc of 0.620). 
Qwen2-VL demonstrates fair detection capability, achieving an Acc of $0.805$ on the UQ split but only $25.0\%$ OP, indicating its limitations in fine-grained classification.
The remaining models perform considerably worse: their Acc values generally stay below 0.55 and their OP scores below 35.0\% on both the UF and UQ splits, in some cases approaching random guessing.
This gap, especially for models such as Qwen2.5-VL that perform well on standard VQA benchmarks, suggests that \dataset poses challenges that are substantially different from those in traditional VQA settings and remains far from being solved by current pretrained VLMs.

\mypara{Performance of Fine-Tuned VLMs}
We then fine-tune four open-source VLMs, including Llava-1.5, Qwen2-VL, Qwen2.5-VL, and Gemma3, as detectors on \dataset, and evaluate them on both the COCO and ADE20K-based splits, each with UQ and UF splits.
Across all settings (refer to~\Cref{fig:performance_overall}), fine-tuning yields very strong image-level detection: Acc values typically lie between $0.97$ and $0.99$ on both datasets, indicating that all models can reliably distinguish edited images from real ones once adapted to our task.

At the level of fine-grained localization, Qwen2.5-VL is consistently the strongest detector.
On COCO, when averaged over the $6$ editing models, it achieves about $0.99$ Acc, $76\%$ RP, and $72\%$ OP on the UQ split, and about $0.98$ Acc, $83\%$ RP, and $80\%$ OP on the UF split.
On ADE20K, its performance further improves: the average RP/OP reaches roughly $86\%$/$84\%$ on UQ and $94\%$/$93\%$ on UF, with Acc close to 0.99 in both cases.
By contrast, LLaVA-1.5 is uniformly the weakest among the $4$ detectors, especially on COCO-UQ where its average OP is around 58\% (Acc $\approx$ $0.94$ and RP $\approx$ $63\%$), though even this represents a substantial gain over the pretrained setting.

Comparing UQ and UF reveals a clear difficulty gap.
On COCO, the overall average OP rises from about $64\%$ on UQ to $77\%$ on UF; on ADE20K, it increases from about $81\%$ to $90\%$.
This drop on UQ is expected, since the UQ split enforces non-redundant target objects and therefore requires models to generalize to unseen entities, making it a more challenging and realistic scenario.
In addition, results on ADE20K are persistently stronger than on COCO. 
For example, the average OP across all models and editors improves from $64\%$ (COCO-UQ) to $81\%$ (ADE20K-UQ), and from $77\%$ (COCO-UF) to $90\%$ (ADE20K-UF).
We attribute this gap in part to the higher resolution and more structured scenes in ADE20K, which make edited regions more visually salient and easier for VLMs to exploit.
Overall, these observations show that fine-tuning enables modern VLMs to reach high accuracy and strong fine-grained localization.

\mypara{Broader Edit Instruction Detection}
In our earlier experiments, the detection targets were primarily the most common and well-developed editing operations in current image editing models, namely object addition and object replacement.
With the more capable open-source editing model Step1X-Edit, we further expand the detection scope to broader editing types, including background change and facial expression change.
For background change, we use the ADE20K, while for facial expression change, we use the FFHQ.
Following the same experimental setup, we evaluate the detectors on the UF-split of the dataset.
\begin{figure}
    \centering
    \includegraphics[width=0.98\linewidth]{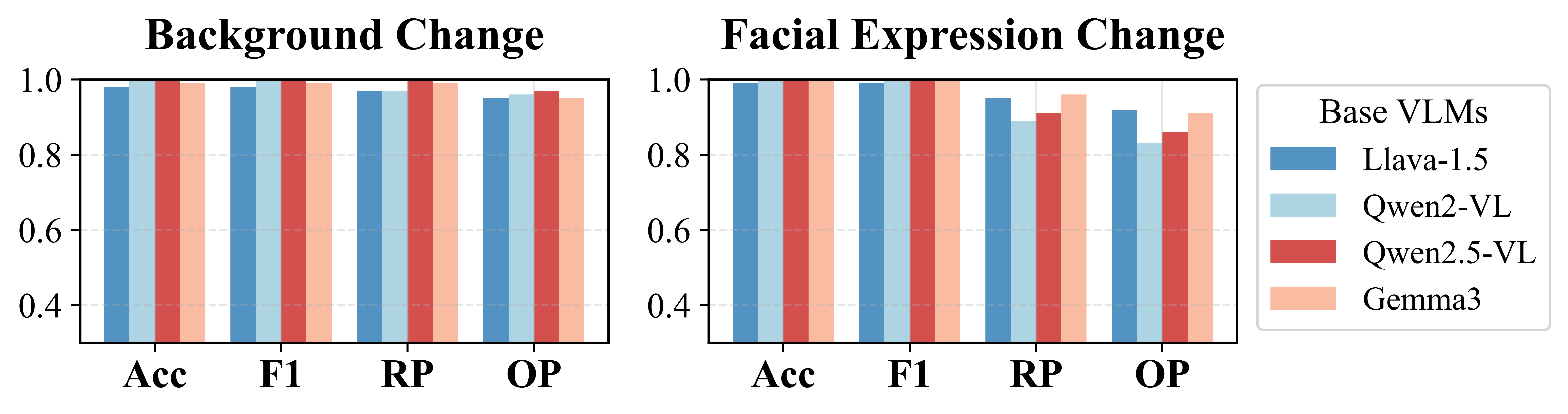}
    \caption{Detection performance on background and face-expression edits.}
    \label{fig:comparison}
\end{figure}
As shown in~\Cref{fig:comparison}, all $4$ VLMs perform substantially better on background-change detection than on object addition and replacement, with Accuracy and F1 almost saturated and both RP and OP close to $1.00$.
For face-expression change, the overall performance is similar to that in the previous settings, though Gemma3 achieves the best localization with an RP of $96\%$ and an OP of $91\%$.
This pattern is likely because background edits modify large regions of the image and thus provide stronger visual cues, whereas face-expression edits only affect a small area and introduce more limited visual changes.

%-----------------------
\subsection{Ablation Study}
%-----------------------

\mypara{Effect of LoRA Rank on Detection Performance}
In our LoRA-based fine-tuning, the rank determines how many additional parameters are introduced, and we vary this rank to examine its impact on edited-image detection performance.

\begin{figure}[!h]
  \centering
  \begin{subfigure}{0.48\columnwidth}
    \centering
    \includegraphics[width=\linewidth]{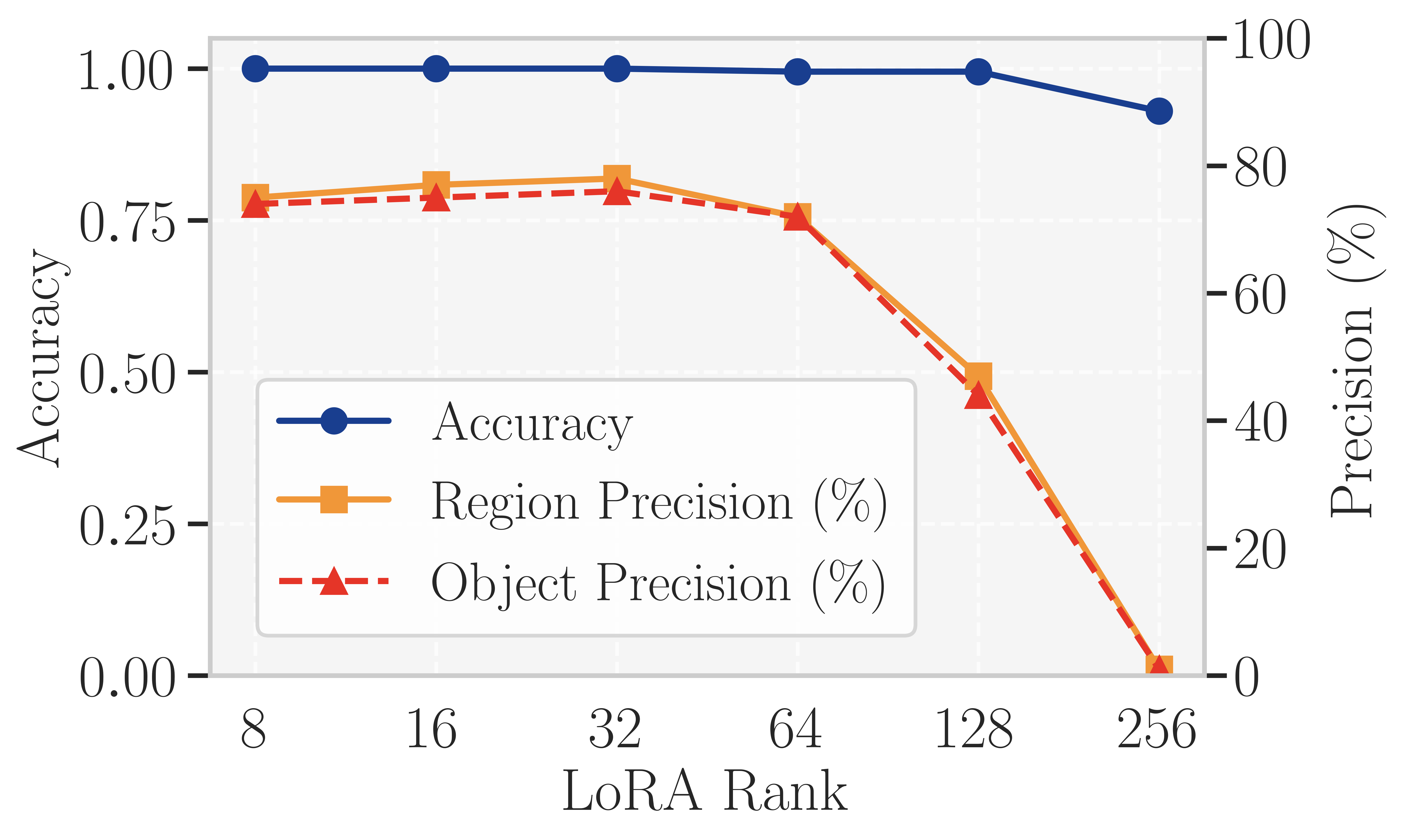}
    \caption{Performance across different rank settings.}
    \label{fig:rank_gemma3}
  \end{subfigure}
  \hfill
  \begin{subfigure}{0.48\columnwidth}
    \centering
    \includegraphics[width=\linewidth]{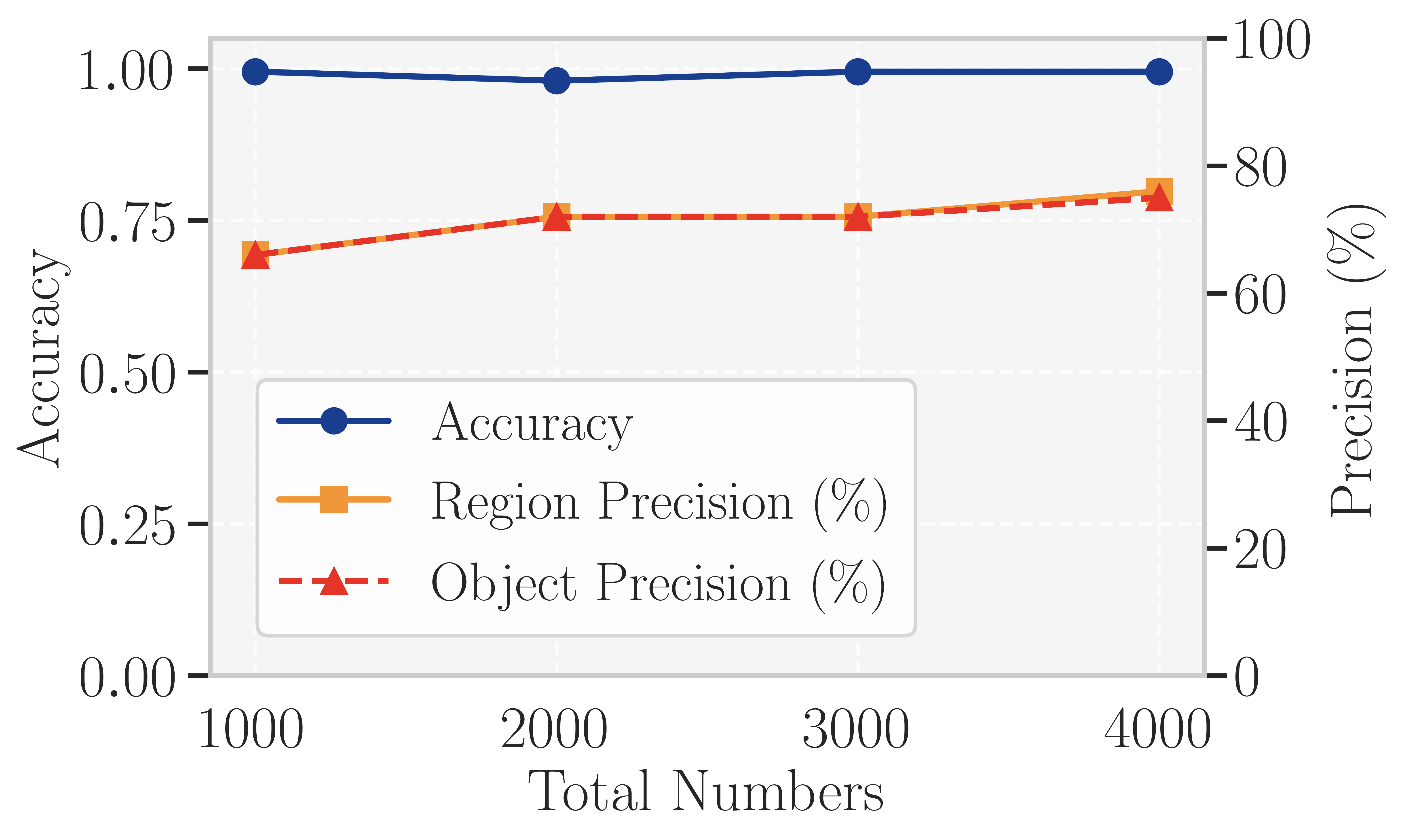}
    \caption{Scaling behavior of training dataset.}
    \label{fig:scaling}
  \end{subfigure}
  
  \caption{Gemma3 detector performance and scaling behavior on Gemini-IG (UF).}
  \label{fig:gemma3_rank_scaling}
\end{figure}

\Cref{fig:rank_gemma3} and \Cref{fig:rank-qwen} show the trends in classification Acc, RP, and OP on the Gemini-IG UF split as the LoRA rank increases. 
For Gemma3 (4B parameters), performance improves with increasing rank and peaks at rank 32 with an Acc of $1.000$, an RP of $78\%$ and an OP of $76\%$. 
Beyond this rank, all three metrics decline.
For Qwen2.5-VL (7B parameters), the best performance occurs at rank 8, with a Region Precision of $81\%$ and an Object Precision of $78\%$. 
These results suggest that different VLMs can have different optimal LoRA ranks, and that increasing the rank beyond this point may even hurt performance of the base VLMs.

\begin{figure}[h!]
    \centering
    \includegraphics[width=1\linewidth]{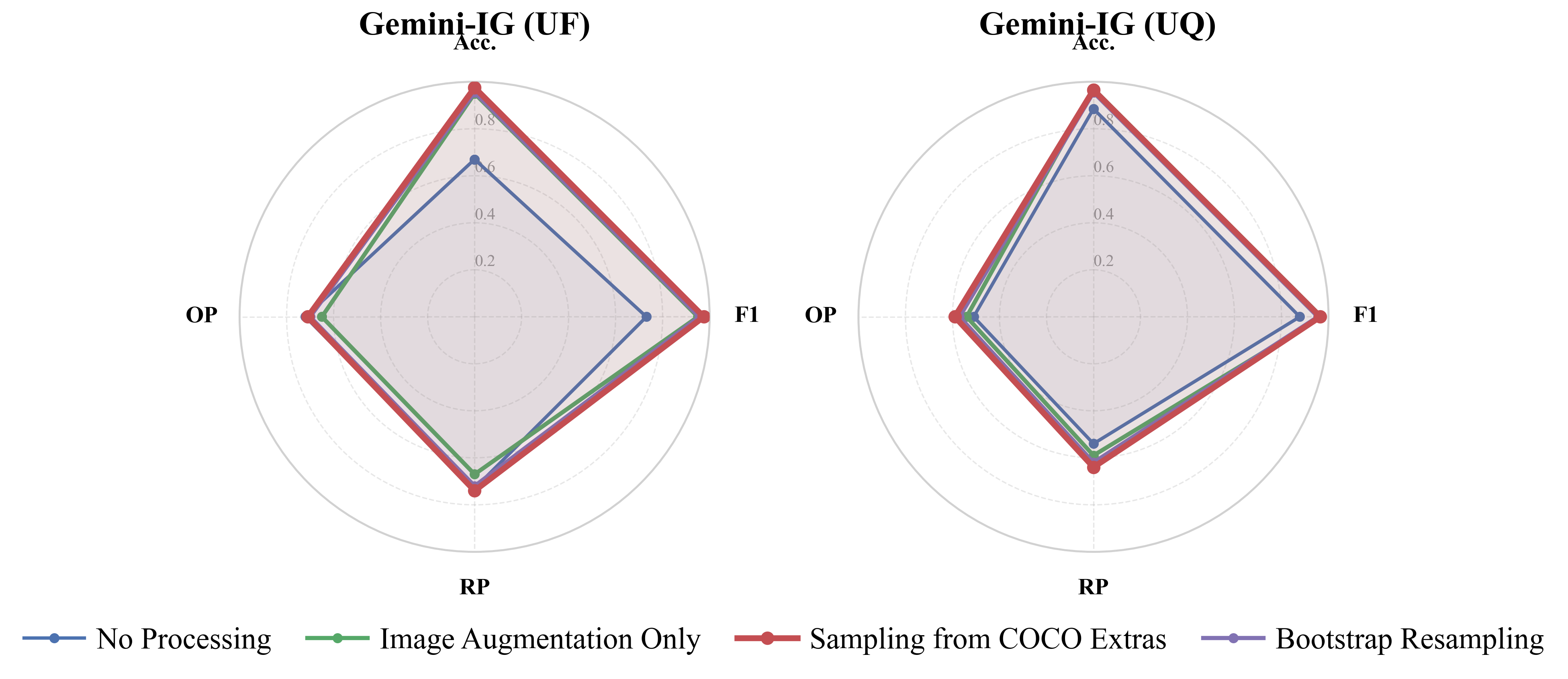}
    \caption{Performance comparison of fine-tuned Llava-1.5 model trained on a 4,000-sample Gemini-IG subset using different data preparation strategies.}
    \label{tab:gemini-llava-all}
\end{figure}

\mypara{Comparison of Different Data Balancing Strategies}
COCO subset of \dataset contains $1,600$ original images, with 100 for testing. 
We now consider a setting where the number of edited training images is fixed at $2,000$ and exceeds the number of original images, and we compare different strategies for balancing the training data.
As shown in~\Cref{tab:gemini-llava-all}, for the No Processing baseline, we train on all available data ($1,500$ original + $2,000$ edited = $3,500$ images), which yields $0.885$ Acc with $51.0\%$ OP on the UQ split and $67.0\%$ Acc with $72.0\%$ OP on the UF split.
To utilize the full $4,000$-image budget and keep original and edited images balanced at $2,000$ each, we expand the original set from $1,500$ to $2,000$ via one of three strategies: 
(1) Image Augmentation Only generates $500$ new samples by applying random rotations, horizontal flips, and center crops to the $1,500$ originals; 
(2) Sampling from COCO Extras draws $500$ images from the remainder of the COCO dataset not used in the original split; 
(3) Bootstrap Resampling samples with replacement from the $1,500$ originals until the set reaches $2,000$ images.

All three balancing strategies clearly outperform the no-processing baseline. Among them, Sampling from COCO Extras works best, pushing UQ performance to $0.97$ Acc / $59\%$ OP (with the highest UQ RP of $64\%$) and UF performance to $0.98$ Acc / $71\%$ OP, while simple augmentation or bootstrap resampling bring smaller but still noticeable gains.

\begin{figure}[h!]
    \centering
    \includegraphics[width=0.90\linewidth]{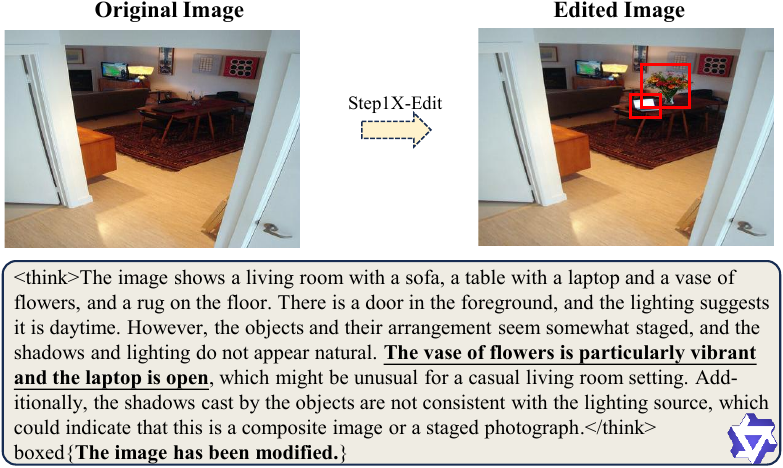}
    \caption{Output example of Qwen2.5-VL (RLVR)}
    \label{fig:rlvr_case}
\end{figure}

\begin{figure*}[!h]
  \centering
  \includegraphics[width=0.8\textwidth]{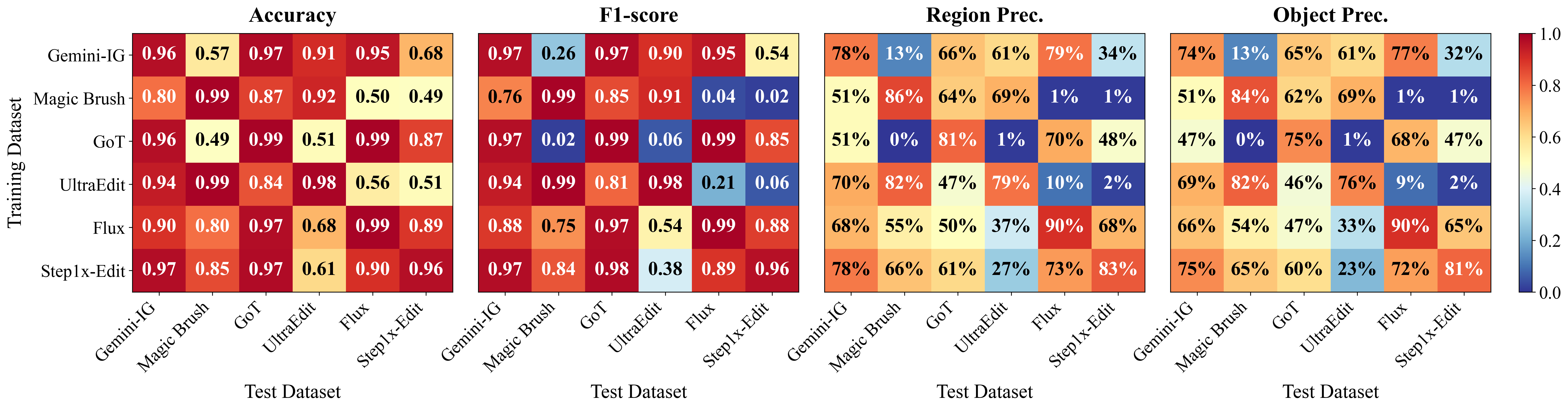}
  \caption{Cross-editors transferability of Qwen2.5-VL under the COCO (UF split).}
  \label{fig:heatmaps_easy}
\end{figure*}

\mypara{Effect of Data Scale}
We evaluate how performance scales by fine-tuning Gemma3 with LoRA on the Gemini-IG subset while varying the training size from $1,000$ to $4,000$ images, keeping a 1:1 ratio between original and edited images.

As shown in~\Cref{fig:scaling}, overall classification accuracy remains nearly $1.00$ across all sample sizes. 
In contrast, both RP and OP improve steadily as the dataset grows. 
RP increases from $66\%$ at $1,000$ images to $76\%$ at $4,000$ images, while OP rises from $66\%$ to $75\%$. 
These findings indicate that although classification accuracy saturates at an early stage, the more detailed metrics continue to benefit from larger training sets.

\mypara{Comparison of RLVR Training}
\Cref{tab:rlvr_training} compares SFT and RLVR when fine-tuning Qwen2.5-VL on the Step1X-Edit UF split.
RLVR yields only modest but consistent gains over SFT: F1 increases from $0.96$ to $0.98$, RP from $83\%$ to $85\%$, and OP from $81\%$ to $84\%$, while Acc remains at $0.97$.
More importantly, RLVR optimizes the model under a verifiable reward that explicitly encourages structured reasoning and localization of the manipulated content.
As illustrated in \Cref{fig:rlvr_case}, the RLVR-trained model provides a natural-language explanation and highlights the suspicious objects and regions.
This makes the detector's behavior more interpretable and offers concrete visual and textual cues that can assist non-expert users in understanding and verifying the detection outcome.

\begin{table}[h!]
\centering
\small
\caption{RLVR vs. SFT on Step1X-Edit based on Qwen2.5-VL.}
\label{tab:rlvr_training}
\begin{tabular}{lcccc}
\toprule
 & Acc & F1  & RP & OP \\
\midrule
SFT   & 0.97 & 0.96 & 83\% & 81\% \\
RLVR  & 0.97
                   & 0.98 \textcolor{nicegreen}{(+0.01)}
                   & 85\%  \textcolor{nicegreen}{(+2)}
                   & 84\%  \textcolor{nicegreen}{(+3)} \\
\bottomrule
\end{tabular}
\end{table}

\mypara{Comparison of Different Vision Backbones}
We evaluate the performance of traditional vision backbones on the edited image detection task using the Gemini-IG dataset (UF split). 
The results are shown in~\Cref{tab:backbone-comparison}.
We compare seven visual backbones in two groups: convolutional networks and transformer-based networks.
The Acc of convolutional networks (ResNet-50, DenseNet-121, MobileNet-V2 and Inception-V3) ranges from $0.86$ to $0.91$. 
MobileNet-V2 achieves the lowest Acc at $0.86$, while DenseNet-121 and Inception-V3 both reach $0.91$. 
Transformer-based backbones (ViT-B/16, ConvNeXt-Base and Swin-B/4W7) exhibit greater variation: ViT-B/16 attains $0.94$, ConvNeXt-Base achieves $0.99$, and Swin-B/4W7 achieves $1.00$. 
In VLMs, the current top performer Gemma3 also reaches $1.00$ Acc (see~\Cref{fig:rank_gemma3}). 
However, despite their strong accuracy, these backbones provide only image-level predictions and lack the fine-grained localization and object descriptions that are crucial for real-world forensic applications.

\mypara{User Study}
To further explore the gap between fine-tuned VLMs and human perception, we design a subjective evaluation questionnaire.
Five images edited by the Flux model and five original images are randomly selected and shuffled.
Using the questionnaire shown in~\Cref{fig:Questionnaire}, we collect $42$ valid responses from non-expert volunteers online, and the results are manually analyzed.
\begin{table}[h!]
\centering
\small
\caption{Results of the user study.}
\label{tab:user_study}
\begin{tabular}{lcccc}
\toprule
 & Acc & F1  & RP & OP \\
\midrule
Volunteers  & 0.62 & 0.60  & 45.2\% & 33.8\% \\
\bottomrule
\end{tabular}
\end{table}
As shown in \Cref{tab:user_study}, the volunteers achieve an Acc of only $0.62$, which is significantly lower than the fine-tuned Qwen2.5-VL ($0.99$). 
The error distribution in~\Cref{fig:error_analysis} indicates that participants frequently miss actual edits, often making multiple mistakes on edited images (e.g., 2, 3, or even 5 errors), and also mislabel a non-negligible number of original images as edited.
Their RP for locating edited areas is only $45.2\%$, further highlighting the limitations of non-expert humans in AI-edited image detection and the need for robust detectors.

%-----------------------
\subsection{Zero-Shot Transferability}
%-----------------------

Here, we investigate the detectors' generalization to unseen editing scenarios without further fine-tuning.

\mypara{Transferability on Different Editor Datasets}
We evaluate the zero-shot transferability of Qwen2.5-VL by training on $1$ editor dataset and testing on the remaining $5$ without further fine-tuning, under both the UF and UQ splits, as shown in~\Cref{fig:heatmaps_easy,fig:heatmaps_hard}.

Across all cross-dataset settings, the OP drops noticeably: the average off-diagonal OP is about $45\%$ on UF and $34\%$ on UQ, even though the corresponding accuracies typically remain around 0.8.
This indicates that the detector often preserves a reasonable binary decision, but struggles to localize the manipulated object when the editing style differs from the training domain.
The choice of training dataset strongly influences transfer behavior.
Detectors trained on Step1X-Edit exhibit the most robust cross-dataset performance, with average off-diagonal OP of roughly $59\%$ on UF and $47\%$ on UQ (e.g., $75\%$ and $69\%$ OP when transferred to Gemini-IG, and $72\%$ and $54\%$ when transferred to Flux).
Flux-trained detectors also generalize relatively well.
In contrast, MagicBrush-trained detectors tend to overfit to their own artifacts: their cross-dataset OP can collapse to single digits in several cases (e.g., around $1\%$ when transferred to Flux on UF and UQ), although they still retain comparatively high OP when transferred to UltraEdit, suggesting that those two editing pipelines share more similar visual characteristics.
A similar pattern appears between Flux and Step1X-Edit, which achieve mutually high OP when transferred to each other.

Overall, these results show that detectors trained on a single editing style do not reliably generalize to unseen generators, especially for fine-grained localization and object identification.
Robust open-world edited image detection requires joint training on diverse editing datasets rather than relying on a single-source supervisor.

\begin{figure}[h!]
  \centering
  \includegraphics[width=0.8\linewidth]{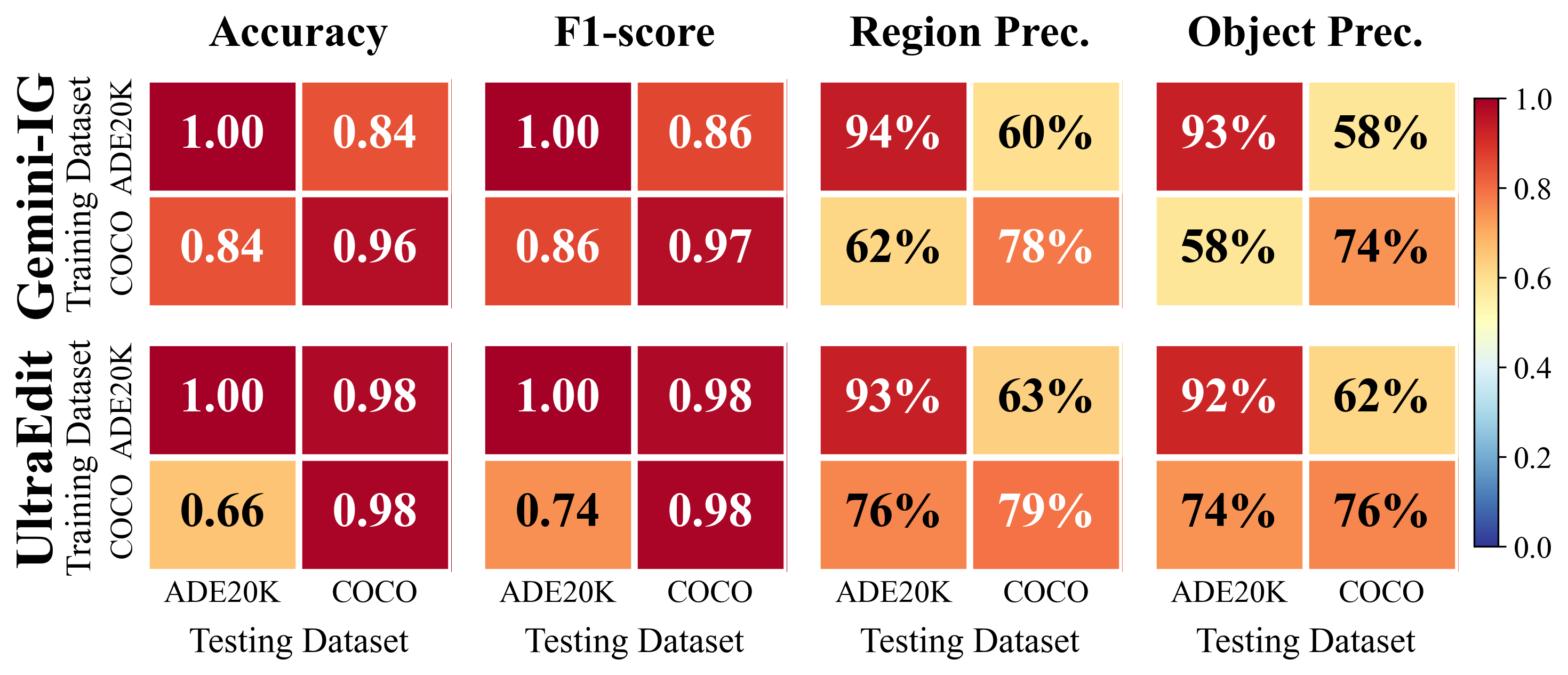}
  \caption{Cross-dataset transferability of Qwen2.5-VL between COCO and ADE20K as original-image sources on Gemini-IG and UltraEdit (UF split).}
  \label{fig:transfer_coco_ade}
\end{figure}

\mypara{Transferability across Original Image Datasets}
We further examine cross-dataset transfer between COCO and ADE20K as original-image sources using Qwen2.5-VL with LoRA on Gemini-IG and UltraEdit (\Cref{fig:transfer_coco_ade}).
Overall, both datasets provide reasonably transferable supervision, but with a clear drop compared to in-domain performance: when training on one dataset and testing on the other, Acc typically remains above $0.80$ while OP falls to about $58$-$76\%$, which is $15$-$30$ points lower than the corresponding in-domain values.
For Gemini-IG, the two directions (COCO$\rightarrow$ADE20K and ADE20K$\rightarrow$COCO) behave similarly, whereas for UltraEdit they show a trade-off, with ADE20K-trained detectors preserving higher Acc on COCO and COCO-trained detectors retaining stronger OP on ADE20K.
These results suggest that cross-dataset generalization is feasible but non-trivial, and depends jointly on the original-image distribution and the editor.

\begin{figure}[!h]
  \centering
 \includegraphics[width=0.8\linewidth]{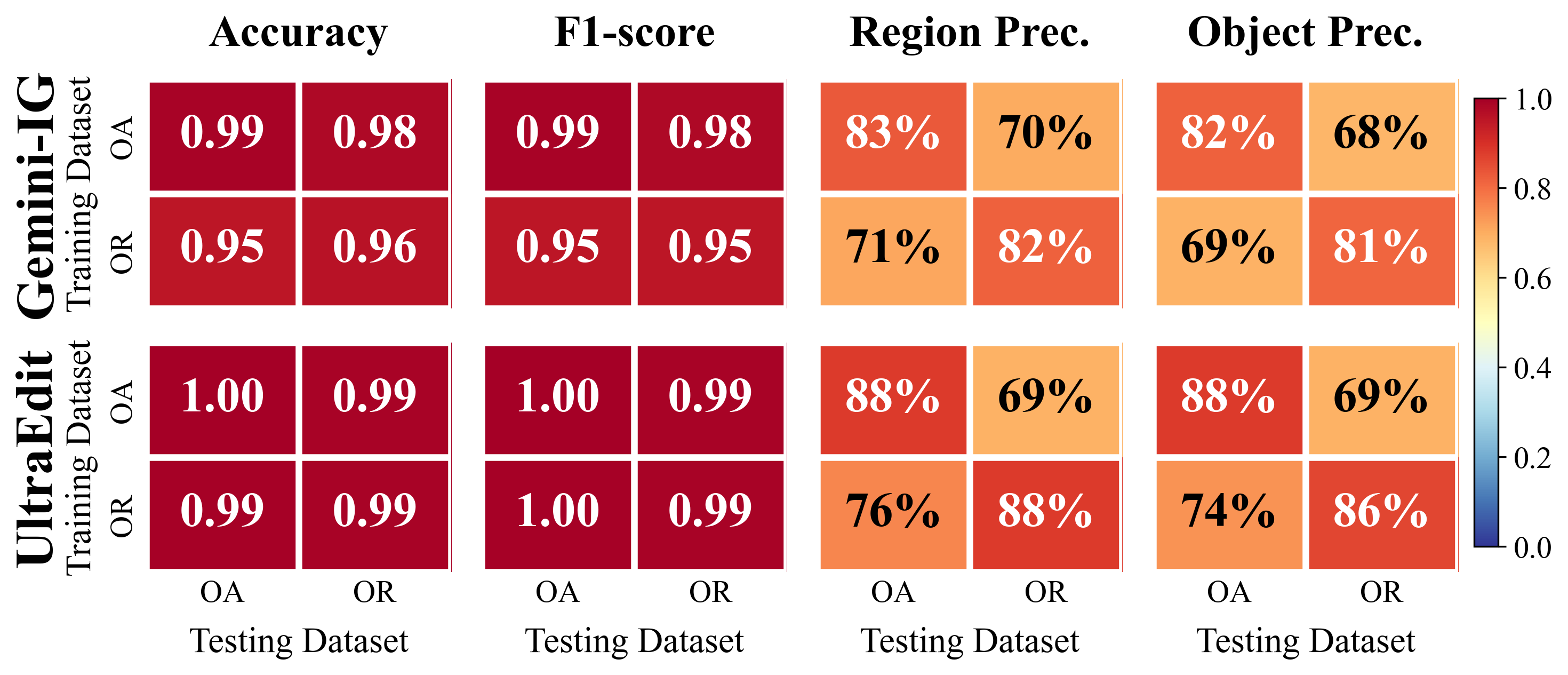}
  \caption{Cross-task transferability of Qwen2.5-VL detector between Object Addition (OA) and Object Replacement (OR) on Gemini-IG and UltraEdit (UF split).}
  \label{fig:transfer_overall}
\end{figure}

\mypara{Transferability on Different Editing Tasks}
We further examine cross-task transfer between OA and OR using Qwen2.5-VL with LoRA on the Gemini-IG and UltraEdit UF splits. For each dataset and task, we train on $2,000$ images and test on 200, with a 1:1 ratio of original to edited images.
As shown in~\Cref{fig:transfer_overall}, the OA- and OR-trained detectors retain high performance when transferred to the other task: on Gemini-IG, OP drops only from $82\%$ to $68$-$69\%$, and on UltraEdit from $88\%$ to $69$-$74\%$, while Acc remains around $0.95$-$1.00$ in all cases.
These results suggest that OA and OR share substantial structure, and that detectors trained on one of these tasks can generalize well to the other without additional fine-tuning.

%-----------------------
\section{Future Directions}
\label{sec:future}
%-----------------------

In this work, we conduct an empirical study of whether VLMs can detect and localize fine-grained AI-edited images, and we introduce \dataset as a benchmark for this problem.
Building on these results, we highlight several promising directions for future research:
\begin{itemize}[left=2pt]
    \item \textbf{Data Filtering Strategies.}
    In this work, we deliberately refrain from filtering the training data, and show that fine-tuned detectors can already achieve strong performance. However, we observe that some editing outputs deviate from the original instructions or modify unintended objects, which introduces noise.
    It would be valuable to explore automated data selection and filtering strategies, for example, using VLM-based judges or methods like LIMA~\cite{DBLP:conf/nips/ZhouLX0SMMEYYZG23} for dataset curation, to construct higher-quality or curriculum-style training subsets and further improve detection performance.

    \item \textbf{Enhancing Transferability.}
    Our experiments reveal clear gaps in transferability across editors and datasets.
    A key future direction is to build detectors that generalize better in open-world settings, for example by jointly training on multiple editing datasets or by using continual learning that helps detectors adapt to new editors and domains with minimal supervision.

    \item \textbf{Richer Training Objectives and Reward Design.}
    We only make an initial attempt with RLVR-based training. In future work, it would be interesting to investigate broader families of RLVR methods, such as DAPO~\cite{DBLP:journals/corr/abs-2503-14476} and GSPO~\cite{DBLP:journals/corr/abs-2507-18071} and to design more expressive reward models that capture not only correctness but also the granularity of the provided explanations.
\end{itemize}
%-----------------------
\section{Conclusion}
%-----------------------

We conduct a detailed empirical study of whether VLMs can detect and localize fine-grained AI-edited images, and we introduce \dataset, a large-scale, fully automatically constructed benchmark specifically designed for this task.
By fine-tuning several open-source VLMs, we show that they can achieve high accuracy in both binary edited-image detection and fine-grained localization of edited objects, substantially outperforming their pretrained counterparts.
We further explore RLVR training in this task, which yields only modest numerical gains but highlights the potential to improve detector interpretability.
Our experiments also reveal non-trivial patterns of transfer across editors, original-image datasets, and editing tasks, highlighting both the promise and the current limitations of VLM-based detectors in open-world settings.
We hope that our work will serve as a foundation for future work on more robust, interpretable, and socially aware image tampering detection.

\bibliographystyle{plain}
\bibliography{ref}

\appendix
\renewcommand{\thefigure}{A\arabic{figure}}
\setcounter{figure}{0}

%-----------------------
\section{Result of Classification in Hive Moderation}
\label{sec:hive}
%-----------------------

\begin{figure}[h!]
    \centering
    \includegraphics[width=0.85\linewidth]{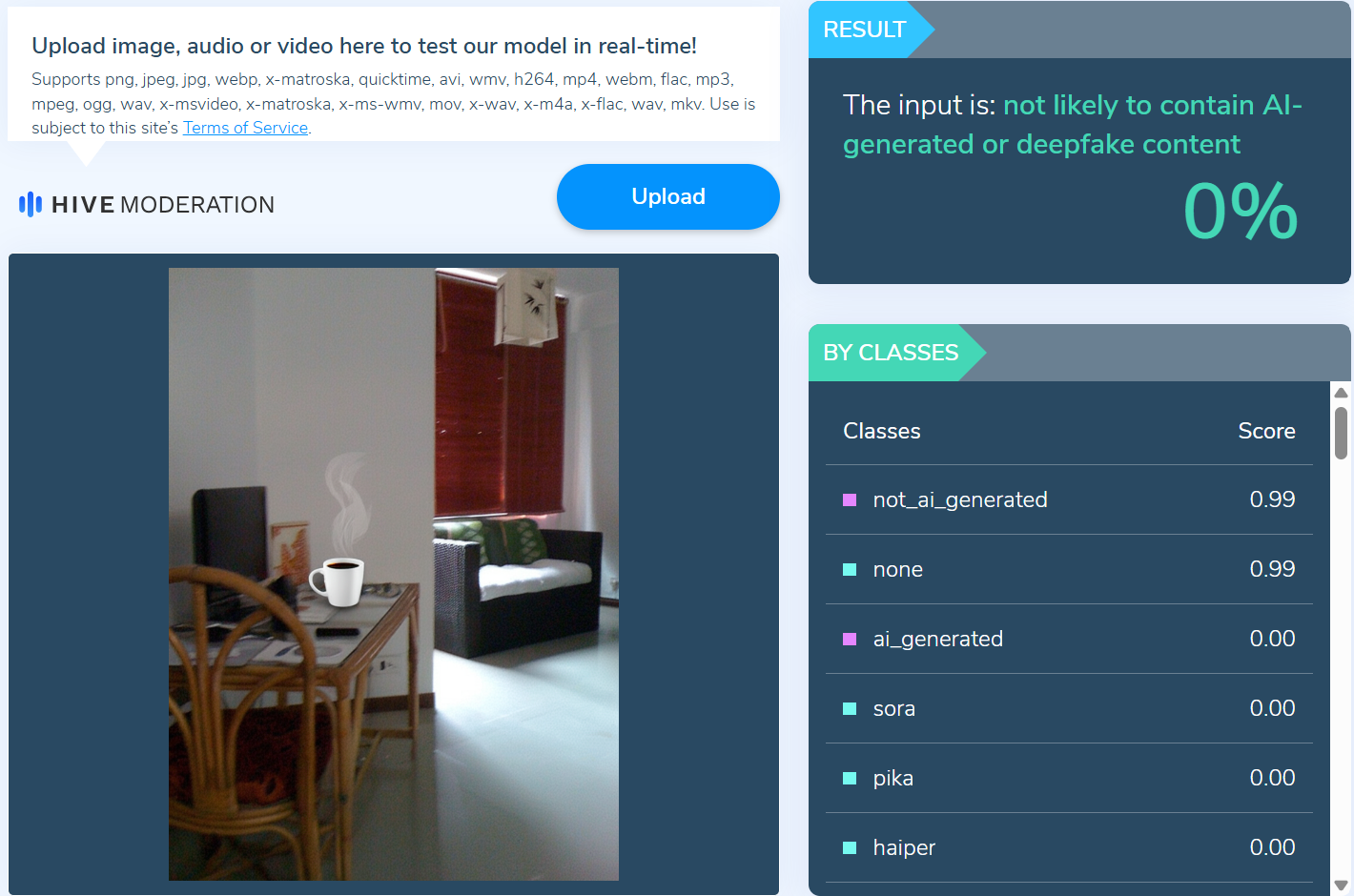}
    \caption{Detection of an edited image using Hive Moderation (ground truth: \textbf{The coffee cup in the image has been modified.})}
    \label{fig:hive}
\end{figure}

Hive Moderation is a commercial fake image detector. We conduct a manual test using 100 edited images from the Gemini-IG Easy test set. We observe that for some edited images, the AI likelihood reported by Hive Moderation is close to zero, as illustrated in~\Cref{fig:hive}. 
Overall, only 55 out of 100 edited images are successfully detected. This result indicates that detectors trained primarily on fully generated images lack robustness when applied to partially edited content.

%-----------------------
\section{Task Template}
%-----------------------

%-----------------------
\subsection{Editing Instruction Creation Template}
\label{sec:creation_template}
%-----------------------

% ---------------box------------------
\newcommand{\gptchat}[1]{%
  \noindent
  \textbf{\faRobot~\textcolor{orange!60!black}{VLM:}} \vspace{--0.8mm}
  \begin{tcolorbox}[colback=orange!6!white, colframe=white, boxrule=0pt, sharp corners=south,
    left=0.8mm, right=0.8mm, top=0.5mm, bottom=0.5mm, boxsep=1pt]
     {\small #1}
  \end{tcolorbox}
}
\newtcolorbox{chatbox}[2][]{
  colframe=orange!60!black,
  colback=orange!3!white,
  fonttitle=\bfseries,
  title=#2,
  enhanced,
  sharp corners=south,
  rounded corners=northwest,
  boxrule=0.6mm,
  width=\linewidth,
  left=3mm,
  right=3mm,
  top=1mm,
  bottom=1mm,
  boxsep=2pt,
  #1
}
\begin{chatbox}[label=chat:creation]{Editing Instruction Creation Template}
You are an expert visual instruction generator specialized in image editing tasks.

Task: Advanced Image Modification Instruction Generation

\textbf{Description:}
Given a COCO dataset object label corresponding to an object in the image, along with the image and a modification goal (e.g., object addition), generate the following four outputs:

1. \textbf{Object Caption:}
   Produce a natural language caption describing the object (e.g., "A woman in the forest wearing a shirt with a drawing.").

2. \textbf{Brief Modification Instruction:}
   Provide a concise, imperative modification instruction (e.g., "Add a fighter jet flying nearby.").

3. \textbf{Symbolic Modification:}
   Output a succinct representation indicating only the addition or removal of objects. Use `+` for additions and `-` for removals.
   - For additions: "+fighter jet"
   - For removals: "-fighter jet"
   - For replacements: "-woman +man"

---

\textbf{Input:}
- \textbf{COCO Object Label:} \{coco\_label\}
- \textbf{Image:} [Provided image]
- \textbf{Modification Goal:} \{object addition/object replacement\}

---

\textbf{Output (in valid JSON format):}
\{
  "object\_caption": "",
  "brief\_modification\_instruction": "",
  "symbolic\_modification": ""
\}

\end{chatbox}
% ---------------box------------------

%-----------------------
\subsection{Edited Image Detection Template}
\label{sec:edited_template}
%-----------------------

% ---------------box------------------
\begin{chatbox}[label=chat:label]{Edited Image Detection Template}
\textbf{Task Description:}

You are a visual analysis assistant. Your task is to examine a given image and determine whether any object in the image has been digitally modified or manipulated. Pay close attention to subtle inconsistencies in lighting, shadows, textures, edges, perspective, or logical composition. Carefully analyze these visual cues before making a judgment.

\textbf{Instructions:}

1. Provide a \textbf{detailed explanation} of your reasoning.

2. Then, based on your analysis, provide a \textbf{final result} in one of the following two formats:
   - If something has been edited:  
     \textbf{"The {thing} in the image has been modified."} 
     (Replace `{thing}' with a brief, clear description of the modified object, e.g., "the cat," "the sky," "the left-side tree," etc.)
   - If nothing has been edited:  
     \textbf{"Nothing has been modified in this image."}

\textbf{Important Constraints:}

- Your explanation must come before the result statement.

- Do not output both statements; only one final result should appear based on your judgment.

- Be cautious: minor edits may be hard to detect but should still be flagged if visible.
\end{chatbox}

%-----------------------
\section{Interpretability Analysis}
%-----------------------

\begin{figure}[htbp]
    \centering
    \includegraphics[width=0.6\linewidth]{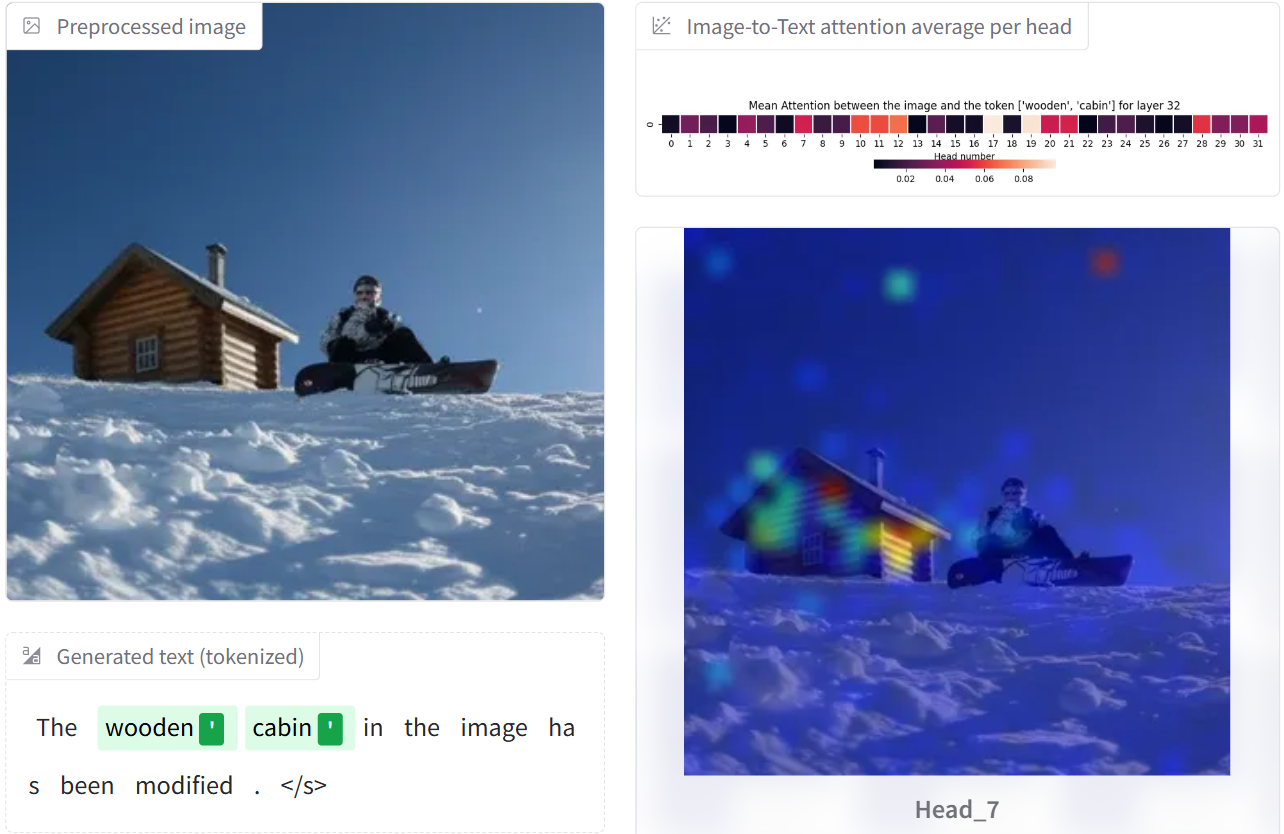}
    \caption{LVLM-Interpret is used to show the model's output for the edited image.}
    \label{fig:attention}
\end{figure}

We perform interpretability analysis on edited images using LVLM-Interpret~\cite{stan2024lvlm}. 
As shown in~\Cref{fig:attention}, the model concentrates its attention on the cabin while generating the detection result. 
This alignment between the predicted modification (“cabin”) and the corresponding visual region demonstrates the reliability and interpretability of the model's output.

%-----------------------
\section{Target Object Analysis}
\label{sec:target}
%-----------------------

To better understand whether GPT-4o tends to propose realistic edits, we analyze the distribution of target objects in the UF splits.
On COCO (UF), it generates $1,600$ object addition instructions with $561$ unique targets (top-1: bicycle, $58$ times) and $1,600$ object replacement instructions with $937$ unique source–target pairs (top-1: baseball bat $\rightarrow$ tennis racket, $16$ times).
On ADE20K (UF), $3,000$ object addition instructions cover $750$ unique targets (top-1: cat, $297$ times), and $3,000$ replacement instructions yield $2,615$ unique pairs (top-1: chandelier $\rightarrow$ pendant light, $15$ times).
For background change (UF), $3,000$ instructions contain $1,145$ unique descriptions (top-1: ``lush green forest'', $100$ times), while for facial expression change (UF), $3,000$ instructions collapse to $62$ unique templates (top-1: ``close his eyes''), reflecting the smaller space of natural facial edits. In the UQ version, all target objects are strictly unique.

\begin{figure}[htbp]
    \centering
    \includegraphics[width=0.85\linewidth]{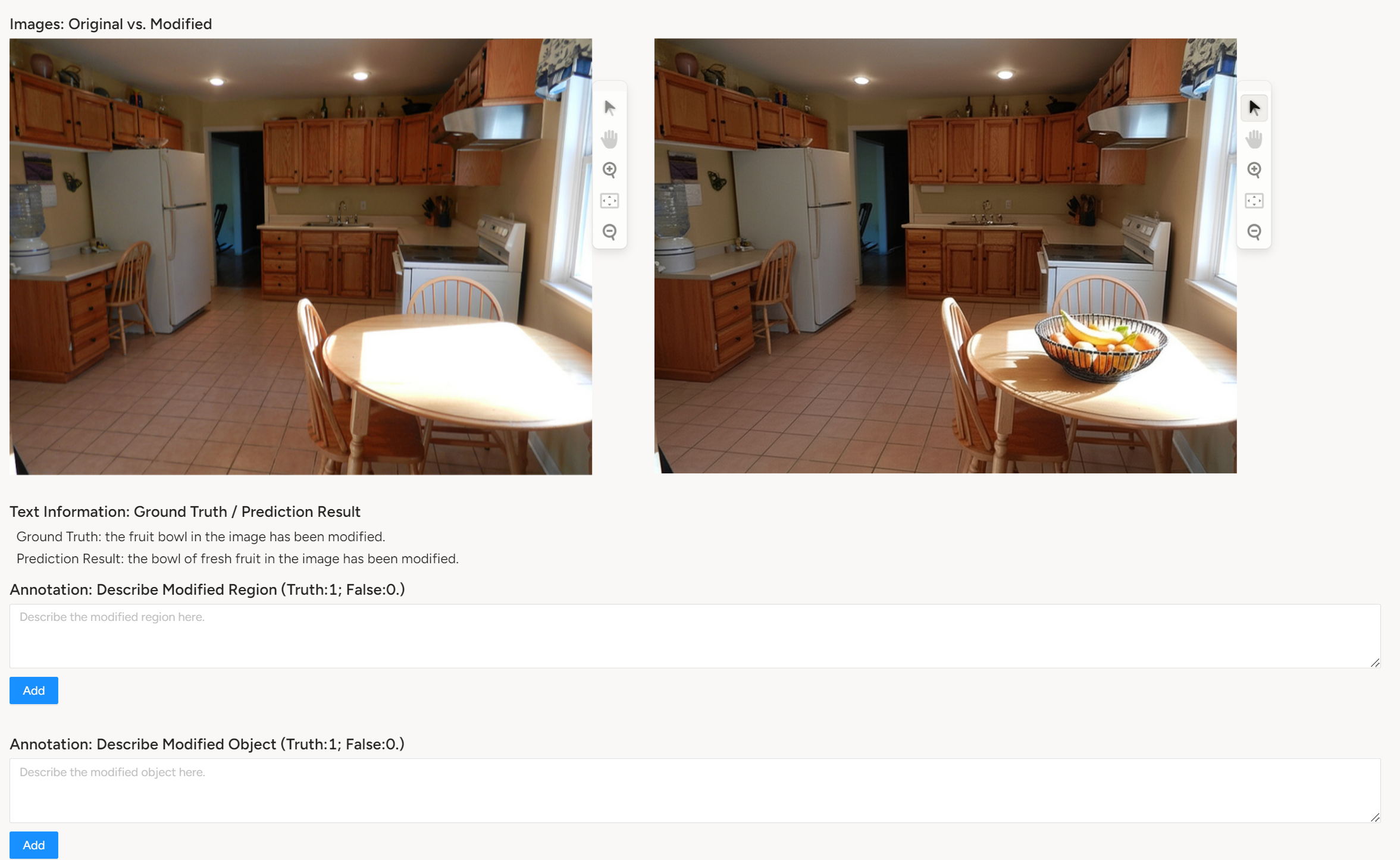}
    \caption{The annotation platform we built using Label Studio.}
    \label{fig:label_studio}
\end{figure}

\begin{figure}[htbp]
  \centering
  \small
  \includegraphics[width=0.7\linewidth]{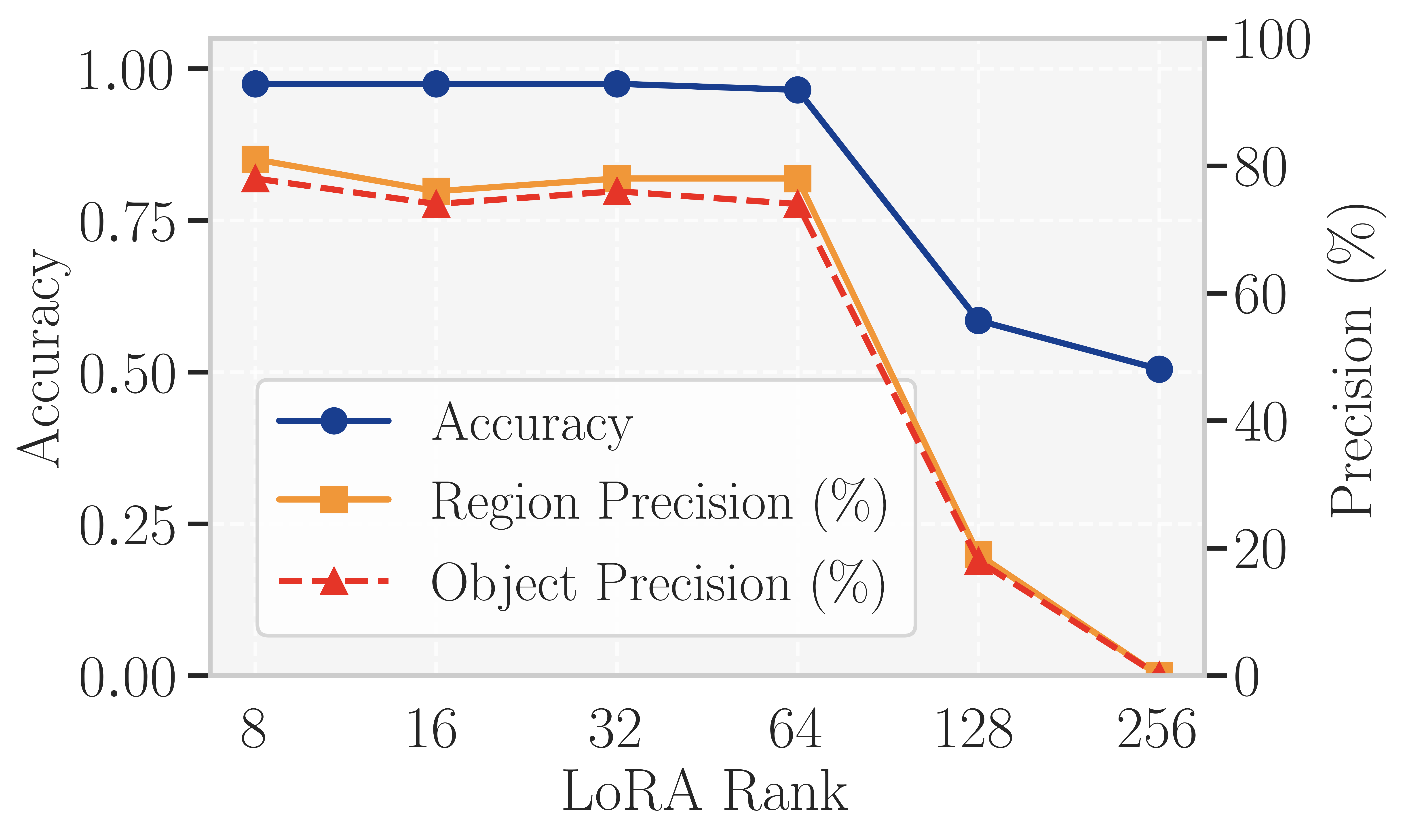}
  \caption{Qwen2.5-VL detector: performance across different rank settings on the Gemini-IG (UF split) dataset.}
  \label{fig:rank-qwen}
\end{figure}

\begin{figure}[htbp]
    \centering
    \begin{subfigure}[b]{0.20\textwidth}
        \centering
        \includegraphics[width=\textwidth]{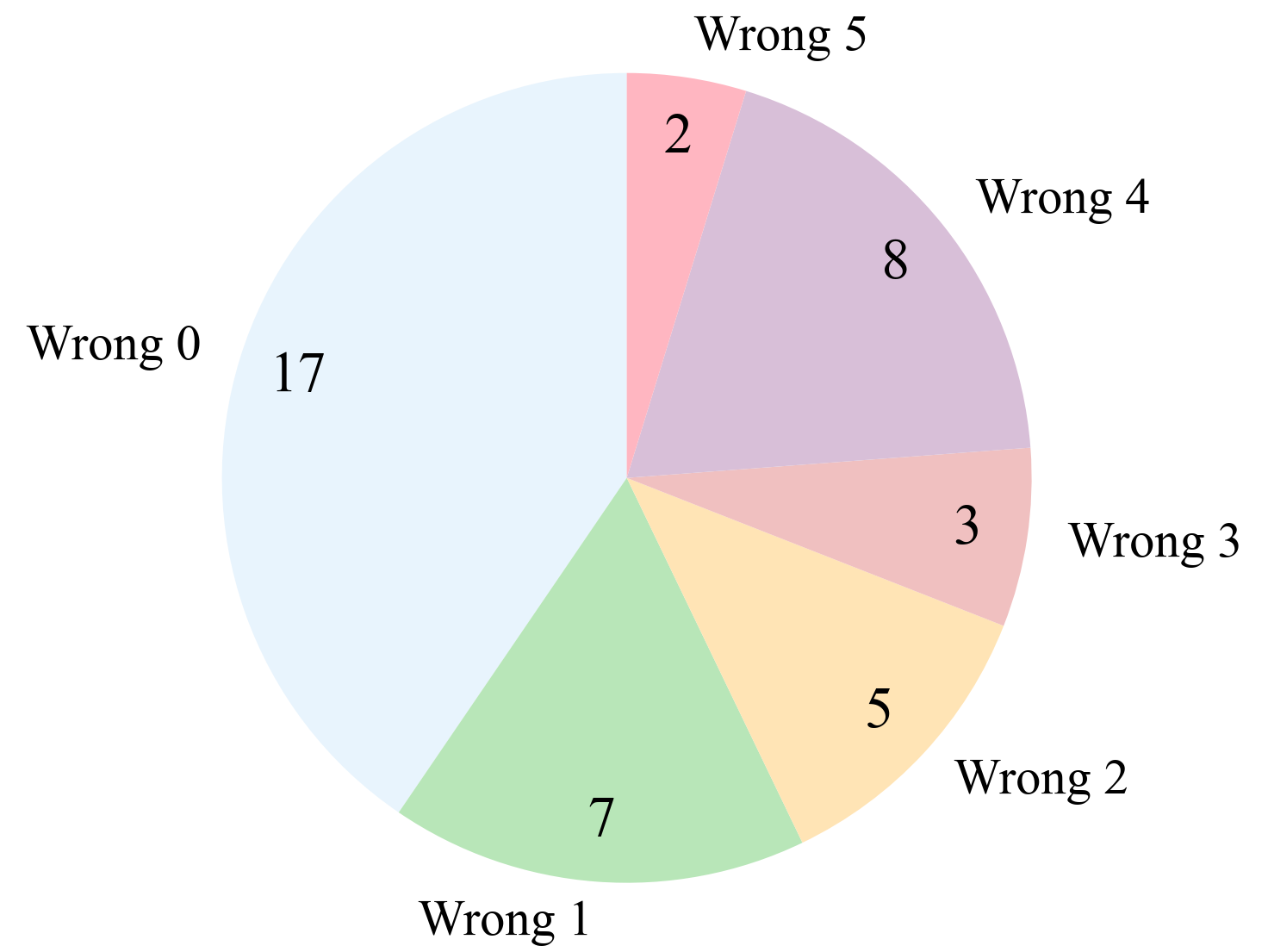}
        \caption{Number of original images misclassified as edited, grouped by how many images are wrongly identified.}
        \label{fig:chart1}
    \end{subfigure}
    \hfill
    \begin{subfigure}[b]{0.20\textwidth}
        \centering
        \includegraphics[width=\textwidth]{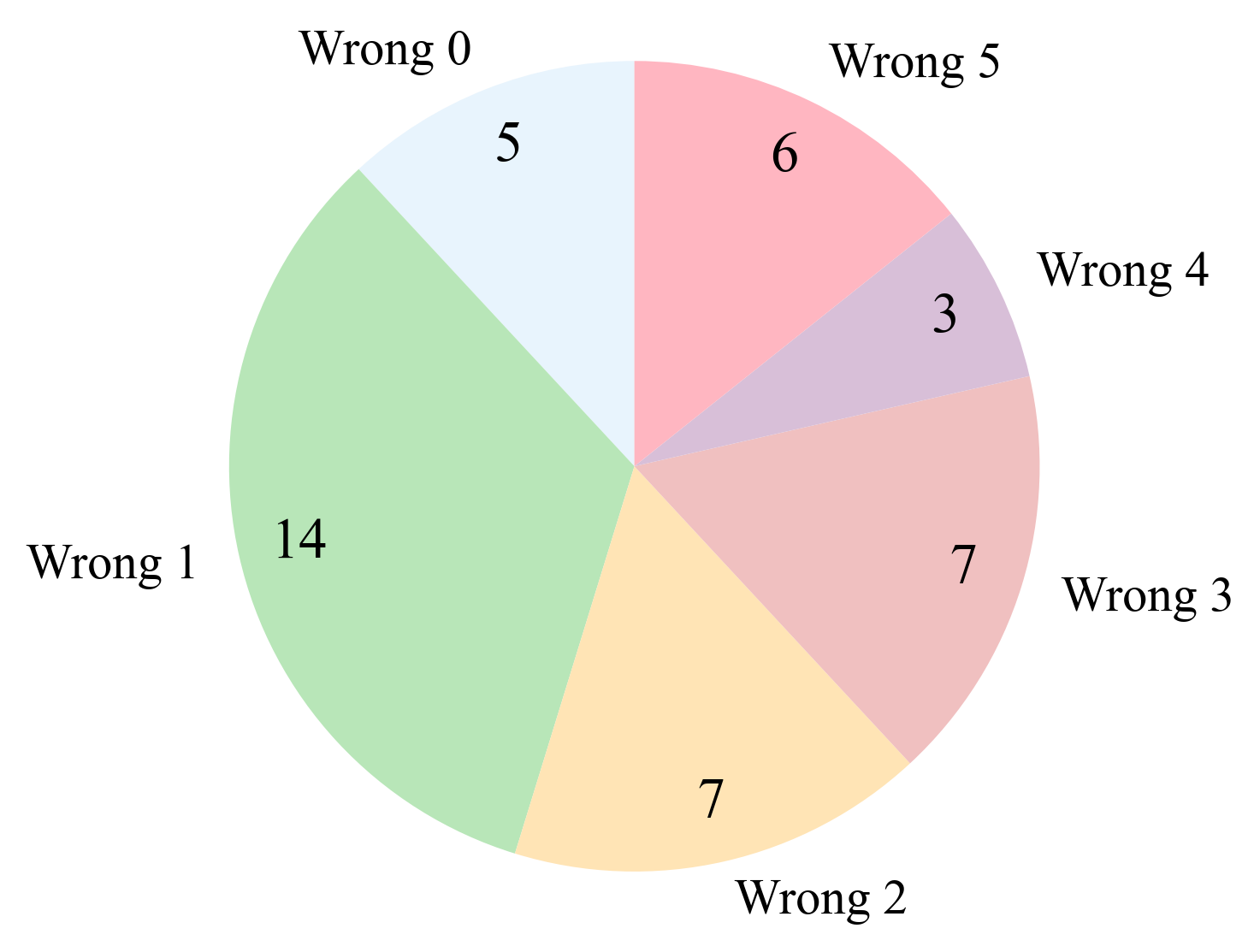}
        \caption{Number of edited images misclassified as original, grouped by how many images are missed.}
        \label{fig:chart2}
    \end{subfigure}
    \caption{Classification error analysis. Labels ``Wrong n'' indicate n incorrect images.}
    \label{fig:error_analysis}
\end{figure}

\begin{figure}[htbp]
    \centering
    \includegraphics[width=0.5\linewidth]{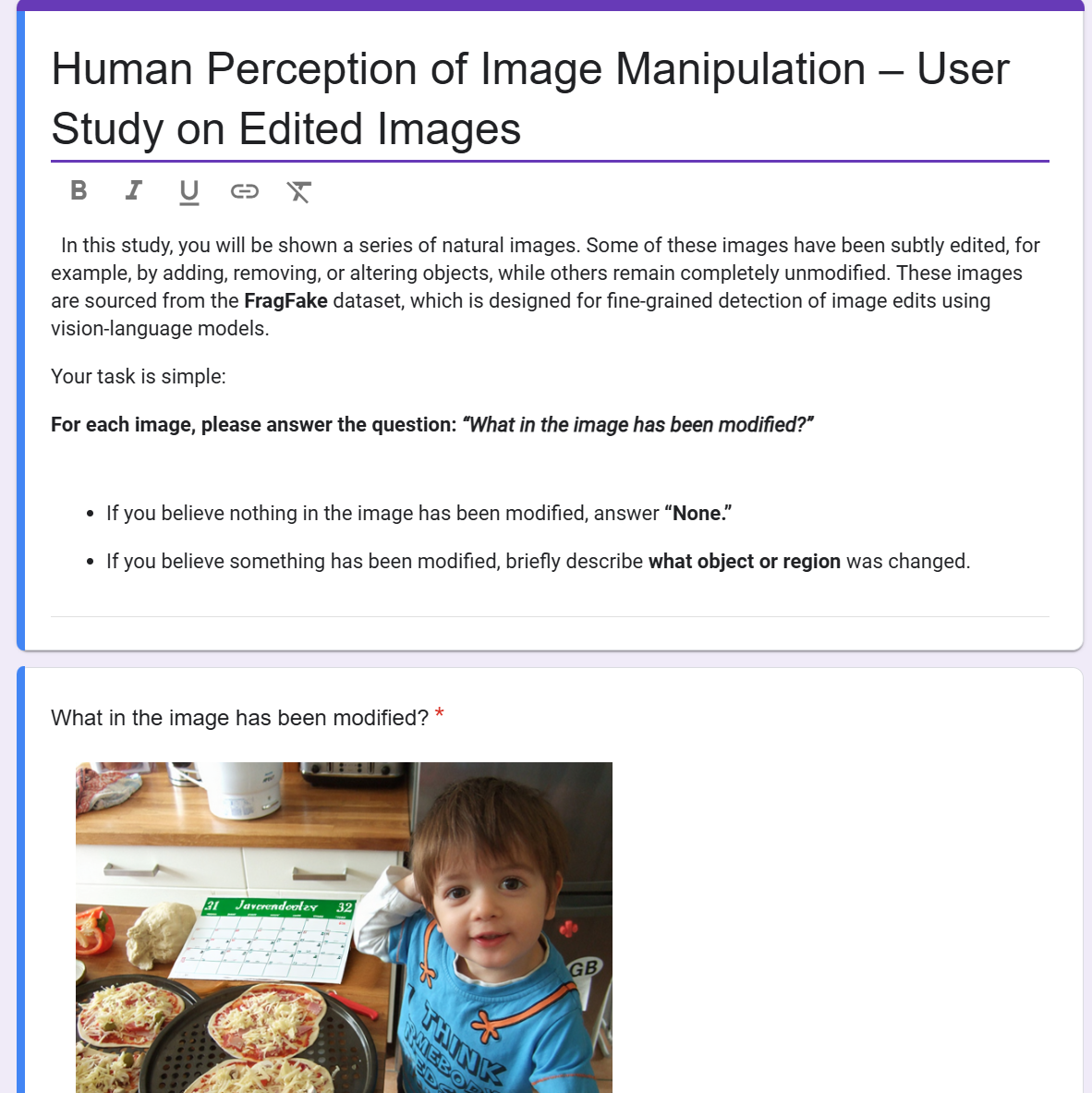}
    \caption{The introductory section of the designed questionnaire, which contains ten questions in total.}
    \label{fig:Questionnaire}
\end{figure}

\begin{table*}[ht]
  \centering
  \small
  \begin{threeparttable}
    \caption{Performance comparison of popular vision backbones on the Gemini-IG (UF split).}
    \label{tab:backbone-comparison}
    \begin{tabular}{l *{7}{C{1.8cm}}}
      \toprule
      \textbf{Metric}
        & \textbf{ResNet-50~\cite{DBLP:conf/cvpr/HeZRS16}}
        & \textbf{DenseNet-121~\cite{DBLP:journals/corr/HuangLW16a}}
        & \textbf{MobileNet-V2~\cite{DBLP:conf/cvpr/SandlerHZZC18}}
        & \textbf{ViT-B/16~\cite{dosovitskiy2021an}}
        & \textbf{Inception-V3~\cite{szegedy2016rethinking}}
        & \textbf{ConvNeXt-Base~\cite{liu2022convnet}}
        & \textbf{Swin-B/4W7~\cite{liu2021swin}} \\
      \midrule
      Accuracy
        & 0.89
        & 0.91
        & 0.86
        & 0.94
        & 0.91
        & 0.99
        & \textbf{1.00} \\
      F1-score
            & 0.89
        & 0.91
        & 0.86
        & 0.94
        & 0.90
        & 0.99
        & \textbf{1.00} \\
      \bottomrule
    \end{tabular}
    \begin{tablenotes}
      \footnotesize
      \item \textbf{Note:} ViT-B/16 = \texttt{vit\_base\_patch16\_224}; Swin-B/4W7 = \texttt{swin\_base\_patch4\_window7\_224}.
    \end{tablenotes}
  \end{threeparttable}
\end{table*}

\begin{figure*}[h!]
    \centering
    \includegraphics[width=0.8\linewidth]{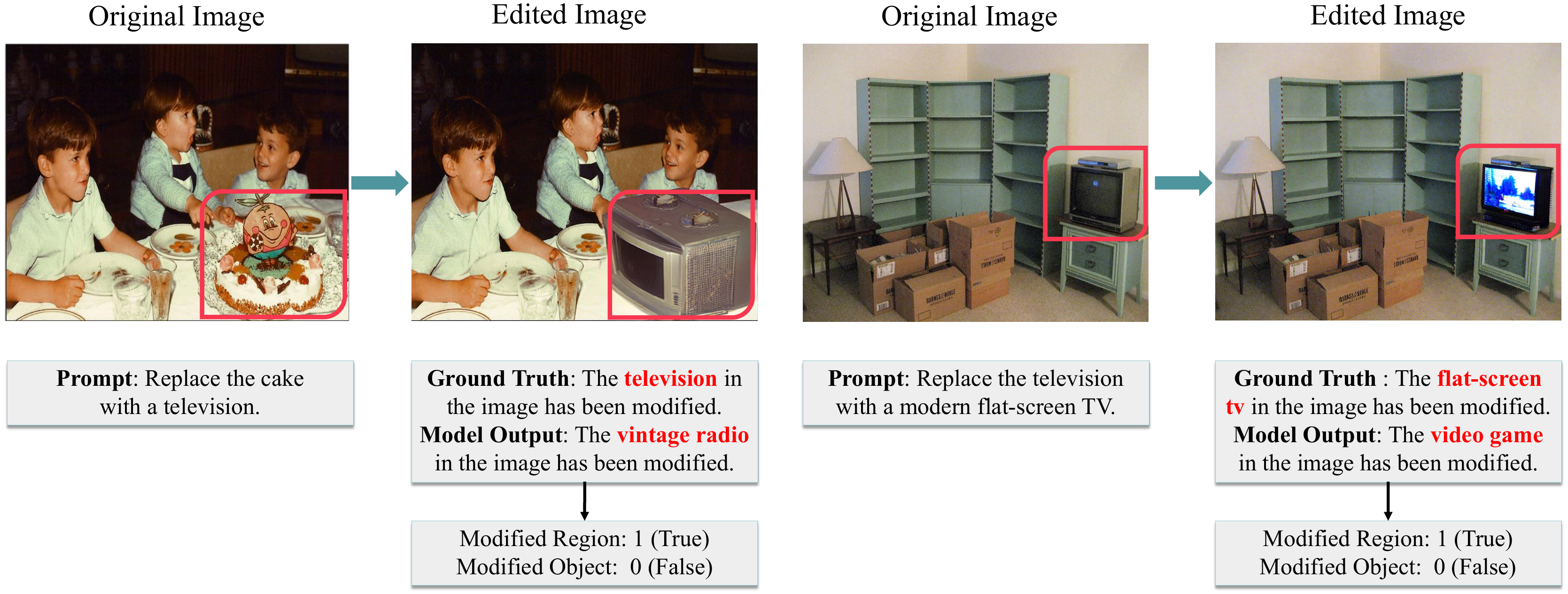}
    \caption{An example of human annotation about the Region Precision.}
    \label{fig:Ground_Truth}
\end{figure*}

\begin{figure*}[h!]
  \centering
  \includegraphics[width=0.8\textwidth]{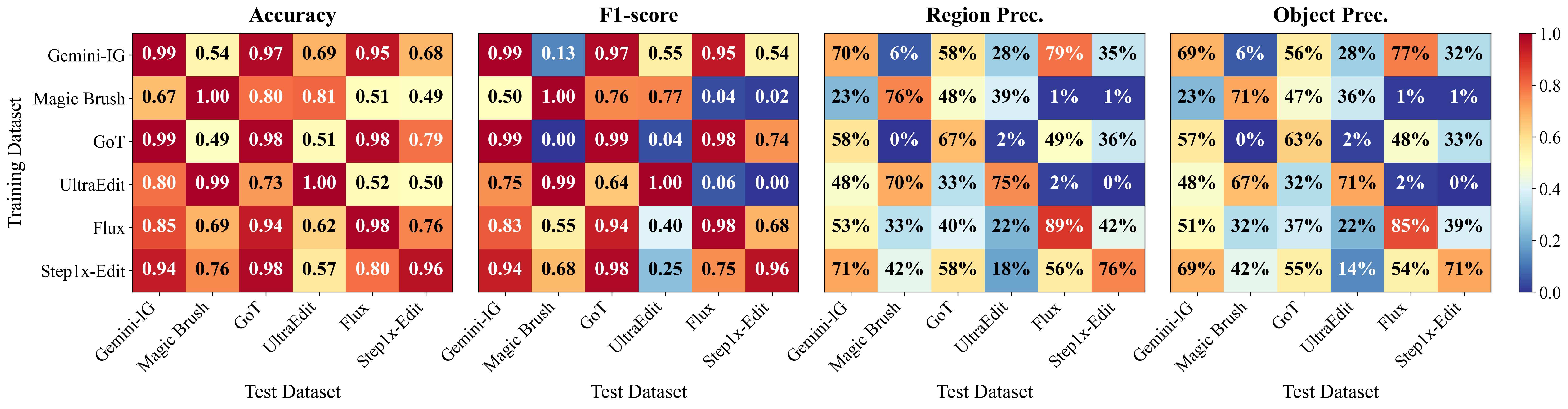}
  \caption{Cross-editors transferability of Qwen2.5-VL under the COCO (UQ split)}
  \label{fig:heatmaps_hard}
\end{figure*}
\clearpage

\end{document}